\documentclass{article}

\PassOptionsToPackage{numbers, sort&compress}{natbib}

\usepackage[main, final]{neurips_2026}
\usepackage[utf8]{inputenc} % allow utf-8 input
\usepackage[T1]{fontenc}    % use 8-bit T1 fonts
\usepackage{hyperref}       % hyperlinks
\usepackage{url}            % simple URL typesetting
\usepackage{booktabs}       % professional-quality tables
\usepackage{amsfonts}       % blackboard math symbols
\usepackage{nicefrac}       % compact symbols for 1/2, etc.
\usepackage{microtype}      % microtypography
\usepackage{xcolor}         % colors
\usepackage{makecell}

\usepackage{xspace}
\usepackage{listings}
\usepackage{adjustbox}
\usepackage{wrapfig}
\usepackage{lipsum}
\usepackage{multirow}
\usepackage{makecell}
\usepackage{pifont}
\usepackage{todonotes}
\usepackage{xcolor}

\makeatletter
\AtBeginDocument{%
  \@ifpackageloaded{algorithm}{%
  }{}%
}
\makeatother
\usepackage{subcaption}

\newcommand{\name}[0]{TokenRouter\xspace}
\newcommand{\xhdr}[1]{{\noindent\bfseries #1}.}

\newcommand{\rom}[1]{\textup{\uppercase\expandafter{\romannumeral#1}}}

\newcounter{observation}

\newcounter{design}

\usepackage[most]{tcolorbox}
\usepackage{colortbl}
\usepackage{pifont}
\usepackage{makecell}

\tcbuselibrary{skins, breakable}   % already loaded if you used it before
\usepackage{tabularx, booktabs}    % flexible table + nice rules

\definecolor{darkred}{rgb}{0.5,0,0}
\definecolor{darkgreen}{rgb}{0,0.5,0}
\definecolor{tahgray}{RGB}{242,242,242}

\newtcolorbox[
  auto counter,
  list inside=examplelist,
]{greenbox}[2][]{
  enhanced,
  title={Example~\thetcbcounter. \textbf{#1}},
  colback=green!5!white,
  colframe=green!50!black,
  colbacktitle=green!60!black,
  coltitle=white,
  #2
}

\newtcolorbox[
  auto counter,
  list inside=examplelist
]{bluebox}[2][]{
  enhanced,
  title={Text~\thetcbcounter. \textbf{#1}},
  colback=cyan!5!white,  % Light blue background
  colframe=cyan!50!black,  % Darker blue frame
  colbacktitle=cyan!75!black,  % Even darker blue title background
  coltitle=white,
  #2
}

\definecolor{TRBg}{HTML}{F8F9FB}
\definecolor{TRBorder}{HTML}{D6D9DE}
\definecolor{TRKeyword}{HTML}{1F4E79}
\definecolor{TRComment}{HTML}{6B7280}
\definecolor{TRString}{HTML}{7A3E9D}

\lstdefinestyle{tokenrouter}{
    language=Python,
    basicstyle=\ttfamily\footnotesize,
    keywordstyle=\bfseries\color{TRKeyword},
    commentstyle=\ttfamily\color{TRComment},
    stringstyle=\color{TRString},
    backgroundcolor=\color{TRBg},
    frame=single,
    framerule=0.35pt,
    rulecolor=\color{TRBorder},
    numbers=none,
    columns=fullflexible,
    keepspaces=true,
    showstringspaces=false,
    breaklines=true,
    breakatwhitespace=true,
    tabsize=2,
    xleftmargin=0.0em,
    xrightmargin=0.0em,
    framexleftmargin=0.8em,
    framexrightmargin=0.8em,
    aboveskip=0.4em,
    belowskip=0.2em
}

\title{TokenRouter: Efficient Serving System for Token-Level LLM Routing}

\author{
    Tianyu Fu\thanks{Equal contribution.} \ $^{1}$,
    Tengxuan Liu$^{* 1}$,
    Ruoxi Wang$^{1}$,
    Yixin Dong$^{2}$,
    \textbf{Yi Ge$^{1}$,
    Yichen You$^{1}$,
    Yu Wang\thanks{Corresponding author: Yu Wang (yu-wang@tsinghua.edu.cn).}} \ $^{1}$
    \\
    \\
    $^1$Tsinghua University\quad
    $^2$Carnegie Mellon University
  }
\begin{document}

\maketitle

\begin{abstract}
Large language model (LLM) routing distributes inference work across different models, advancing the cost-quality Pareto frontier of LLM serving.
While coarse-grained routing at the \emph{session} or \emph{query} level has been widely adopted in production systems, recent algorithmic work shows that fine-grained \emph{token}-level routing can yield substantial efficiency and quality gains.
However, efficiently serving token-level routed inference poses significant challenges to existing systems.
Built on single-LLM assumptions, current systems suffer from severe step desynchronization and frequent batch admission delays under token-level routing, and they also impose high implementation complexity on developers.
To address these challenges, we design \name, an efficient and developer-friendly serving system for token-level routed LLM inference.
\name follows the principle of \emph{request-centric programming, model-centric execution}: developers describe routing logic from the perspective of a single request, while the runtime launches a subserver for each LLM and dispatches requests asynchronously.
Each subserver employs a delayed-batching scheduler, whose optimal hyperparameters are derived from a mathematical throughput model of the system.
Across diverse routing algorithms, workloads, and model pairs, \name achieves 2.01--64.15$\times$ higher decoding throughput than existing systems, substantially advancing the serving efficiency of token-level LLM routing.
Our code is available at \url{https://github.com/thu-nics/TokenRouter}.
\end{abstract}

% Main tex
\begin{figure}[!h]
    \centering
    \begin{subfigure}[b]{0.38\linewidth}
        \centering
        \includegraphics[width=\linewidth]{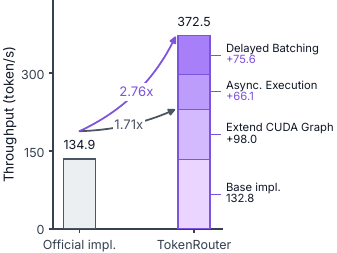}
        \caption{Throughput gain breakdown.}
        \label{fig:gain_breakdown}
    \end{subfigure}%\hfill
    \hspace{10pt}
    \begin{subfigure}[b]{0.53\linewidth}
        \centering
        \includegraphics[width=\linewidth]{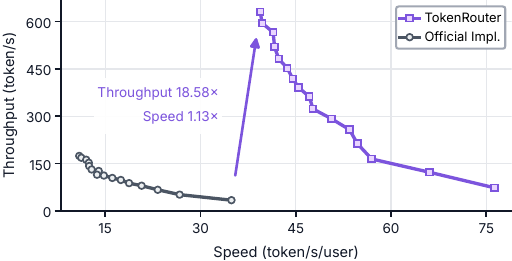}
        \caption{Throughput--speed trade-off.}
        \label{fig:throughput_interactivity}
    \end{subfigure}
    \caption{\name improves serving efficiency for token-level routing.
    (a) At concurrency 8, engineering optimizations yield a 1.71$\times$ throughput gain, and \name achieves 2.76$\times$ overall, relative to the official R2R implementation.
    (b) Throughput--speed trade-off. \name achieves 18.58$\times$ higher throughput than R2R under a stricter SLO.}
    \label{fig:teaser}
\end{figure}

\section{Introduction}

% Background
Large language models (LLMs) have demonstrated strong capabilities across a wide range of applications.
To meet diverse demands, modern LLMs vary in size and domain expertise, showing different latencies and capabilities.
To exploit this diversity and improve the cost-quality Pareto frontier of LLM inference, model routing has emerged as a common serving paradigm.

Coarse-grained routing at \emph{session} or \emph{query} level is already widely used in production LLM serving systems.
For example, commercial systems, such as ChatGPT~\citep{singh2025gpt5} and Cursor~\citep{cursor2026router}, direct each user query to different backend models based on predicted difficulty or topic.
Academic systems such as RouteLLM~\citep{ong2024routellm} also show that query-level routing can achieve strong intelligence at a lower cost with preference data.
However, in these systems, once a request is routed, the response is generated entirely by a single model until the next round.

% What problem
Recent algorithmic work shows that routing at the finer \emph{token} level can unlock benefits that query-level routing cannot reach~\citep{fu2025think,fu2025r2r,huang2026relayllm,jin2024collaborative,liao2026alpharouter,rai2026taro,shen2024learning,xiong2026fusionroute,zheng2025citer}.
% Why important problem
On the efficiency side, token-level routing exploits the sharp variation in generation difficulty within a single query. For example, R2R~\citep{fu2025r2r} can match a 32B model's quality while routing only 5\% of \emph{tokens} to the 32B model and decoding the rest with a 1.5B model, versus routing around 40\% of \emph{queries} to the 32B model.
On the quality side, token-level routing allows models with complementary expertise to collaborate in a single response, surpassing the generation quality of any single LLM~\citep{xiong2026fusionroute, rai2026taro, fu2025cache}.
It also provides smooth control over the entire cost-quality trade-off, rather than the few discrete operating points imposed by coarser routing granularities.

% Why hard problem
However, existing LLM serving systems lack efficient support for token-level routing.
Current systems such as SGLang~\citep{zheng2024sglang} and vLLM~\citep{kwon2023vllm} are designed for single-LLM serving, where all active requests are synchronous at each decoding step.
Token-level routing breaks this assumption by allowing the target LLM to change at every decoding step, creating three main challenges.
First, \textbf{step desynchronization}.
Different LLMs differ widely in per-step latency.
Keeping requests in a single batch synchronized forces every step to wait for the slowest model, leaving the faster models idle.
Second, \textbf{batch admission delay}.
Because token-level routing causes frequent model switches, a routed request often arrives while its target model is still processing a previous batch.
It must then wait to be admitted into the next batch, creating bubbles and fragmenting batches.
Third, \textbf{implementation complexity}.
Current serving systems expose no programming interface for per-step routing decisions.
Implementing token-level routing inside these systems therefore requires extensive modifications to a large codebase.
The routing logic must also be carefully coordinated with tightly coupled features such as continuous batching and prefix caching.

% Solution
To address these challenges, we propose \name, an efficient and developer-friendly serving system for token-level LLM routing.
The key design principle is \textbf{request-centric programming, model-centric execution}: developers describe a routing algorithm by tracing how a single request flows across cooperating LLMs, while \name handles efficient serving through asynchronous model-centric subservers.
This separation hides the system complexity of routing serving from developers and gives the runtime full control to optimize cross-model execution and batch scheduling.
For the user client, \name exposes a single external interface, allowing it to serve as a drop-in replacement for standard single-LLM servers.
Our main contributions are summarized as follows.

\begin{itemize}
    \item \textbf{Request-centric programming interface.}
    We provide a route-send-receive interface that lets developers express diverse token-level routing algorithms by simply tracing how a single request moves across cooperating LLMs.

    \item \textbf{Asynchronous execution for serving optimization.}
    We propose decoupled tri-loop execution that addresses step desynchronization across cooperating LLMs, with a handoff-resume mechanism to simplify inter-model switches.

    \item \textbf{Delayed batching for scheduling optimization.}
    We propose a delayed-batching scheduler that reduces average batch admission latency, with optimal hyperparameters derived from a throughput-based mathematical model.
\end{itemize}

\name brings a high-efficiency LLM serving framework to the token-routing community, turning theoretical benefits into real-world speedups over mature single-LLM serving systems. 
Experiments across different algorithms, workloads, and model pairs show a 2.01--64.15$\times$ throughput boost over existing implementations.

\begin{figure}[t]
    \centering
    \begin{minipage}[b]{0.49\linewidth}
        \centering
        \includegraphics[width=\linewidth]{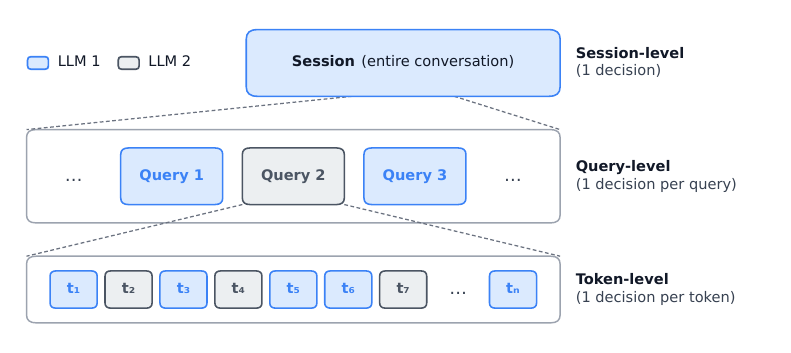}\\
        \small (a) Routing levels in LLM serving.
    \end{minipage}\hfill
    \begin{minipage}[b]{0.49\linewidth}
        \centering
        \includegraphics[width=\linewidth]{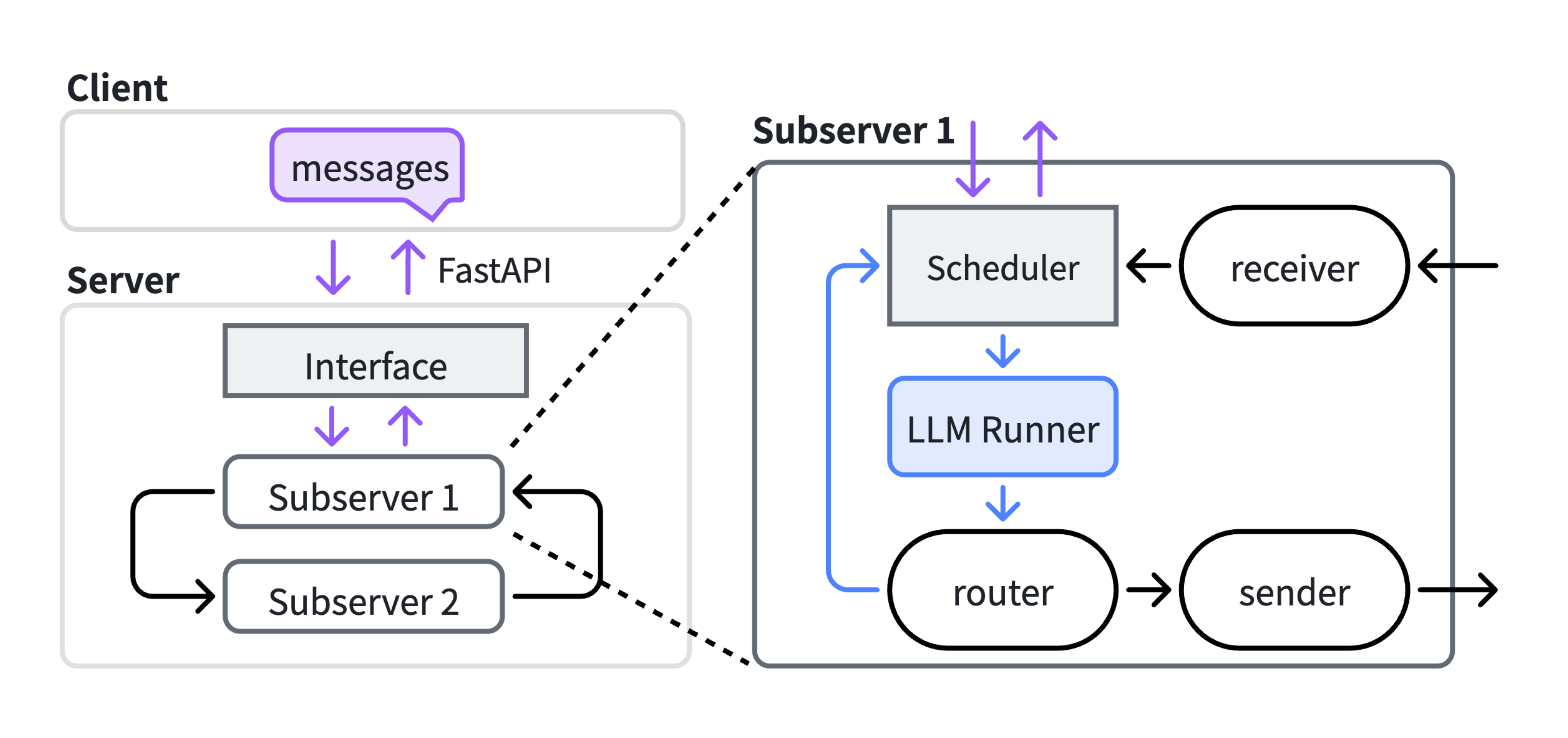}\\
        \small (b) \name system architecture.
    \end{minipage}
    \caption{(a) Routing granularities in LLM serving. Token-level routing allows decisions at each token, whereas session- or query-level routing binds an entire request to one model. (b) \name's model-centric runtime. Each cooperating LLM is hosted by an autonomous subserver that implements three user-defined routing functions (route, send, receive) and runs three decoupled scheduler loops, shown in purple (client-server), blue (decoding), and black (inter-model). The Server Interface exposes a single external endpoint to the client.}
    \label{fig:routing_level}
    \label{fig:architecture}
\end{figure}

\section{Related Work}
\label{sec:related}

\xhdr{Model routing}
Model routing distributes requests across candidate LLMs to improve efficiency or quality.
Most methods operate at the session or query level, choosing models for entire requests or conversation turns.
Efficiency-oriented routers typically send easier requests to cheaper models under a quality constraint~\citep{chen2023frugalgptuselargelanguage,ong2024routellm,ding2024hybridllmcostefficientqualityaware,NEURIPS2024_ecda225c,zhao2024eagleefficienttrainingfreerouter,hu2024routerbenchbenchmarkmultillmrouting,huang2025routerevalcomprehensivebenchmarkrouting}, while quality-oriented routers select the expert best suited to each request and may combine model outputs to surpass any single LLM~\citep{lu2023routingexpertefficientrewardguided,feng2025graphroutergraphbasedrouterllm,wang2024mixtureofagentsenhanceslargelanguage,su2025toolorchestraelevatingintelligenceefficient}.
This coarse granularity lets production serving stacks~\citep{singh2025gpt5} reuse standard single-LLM batching and scheduling.
\name instead targets token-level routing, which requires coordinating model switches within a response.

\xhdr{Token-level routing}
Token-level routing selects a model for each token or short span, enabling collaboration in a single response.
Efficiency-oriented methods invoke larger models for difficult tokens or segments, using trained lightweight routers~\citep{zheng2025citer,fu2025r2r,huang2026relayllm,akhauri2025splitreason,liao2026alpharouter} or heuristic signals such as confidence and entropy~\citep{chen2025rstitch,zeng2026glimprouter}.
Quality-oriented methods combine model expertise by accepting expert tokens~\citep{jin2024collaborative,shen2024learning} or fusing output distributions~\citep{xiong2026fusionroute,rai2026taro}.
Despite different policies, these methods share an execution lifecycle: select next model, transfer the request state, and receive its tokens or distributions to continue generation.
This lifecycle motivates \name's route-send-receive programming interface.
% Beyond algorithmic design, system-level optimization for token-level routing has only been explored in a narrow edge-cloud setting~\citep{she2025token}.

\xhdr{LLM serving systems}
Modern LLM serving systems accelerate inference through continuous batching~\citep{yu2022orca}, paged KV-cache management~\citep{kwon2023vllm}, and efficient structured generation runtimes~\citep{zheng2024sglang}.
These techniques work well for single-model decoding, where each request stays with one model and advances at that model's pace.
Speculative decoding also uses multiple models, but couples draft generation and target verification in a predictable pattern~\citep{leviathan2023fastinferencetransformersspeculative,cai2024medusasimplellminference,li2025eaglespeculativesamplingrequires}.
Token-level routing instead allows data-dependent model switches at each token, leading to irregular model-local arrivals and fragmented batches.
\name addresses these dynamics with a model-centric runtime that hosts each model as an autonomous subserver, preserving continuous batching and prefix caching locally.

% Modern LLM serving systems have made substantial efficiency gains through iteration-level continuous batching~\citep{yu2022orca}, paged KV-cache~\citep{kwon2023vllm}, and efficient structured generation runtimes~\citep{zheng2024sglang}.
% Despite their diversity, these systems share two assumptions. Each request is owned by a single model throughout decoding, and in-flight requests advance step-synchronously at that model's native pace.
% Token-level routing breaks both, since requests change owners at every token and cooperating LLMs run at sharply different speeds.
% Speculative decoding \cite{leviathan2023fastinferencetransformersspeculative, cai2024medusasimplellminference, li2025eaglespeculativesamplingrequires} also uses multiple models, but it couples them through a fixed draft-and-verify pattern with predictable interactions. In contrast, token-level routing induces data-dependent and non-periodic model switches, creating irregular handoffs and fragmented batching.
% \name fills this gap with a model-centric runtime that hosts each LLM as an autonomous subserver coordinated by asynchronous peer messages, while preserving single-LLM optimizations such as continuous batching and prefix caching within each subserver.

\section{Programming Interface}
\label{sec:programming}

To provide a simple and general programming interface, \name abstracts diverse token-level routing algorithms with three main components: \textbf{route}, \textbf{send}, and \textbf{receive}.
These components allow developers to describe when a request should switch models, what state should be transferred, and how returned tokens should be accepted.
Section~\ref{sec:programming/principle} explains the design principle behind this interface, and Section~\ref{sec:programming/interface} presents the concrete API, followed by an example in Section~\ref{sec:programming/example}.

\subsection{Principle: Sequential Request-centric Programming}
\label{sec:programming/principle}

\name is designed to support diverse token-level routing algorithms while keeping their implementation simple and intuitive.
Although \name executes requests through an asynchronous, model-centric runtime, developers should not need to reason about low-level batching, scheduling, or asynchronous execution.
Instead, they should only describe a routing algorithm by tracing how a single request moves among cooperating LLMs.

Therefore, we revisit the common lifecycle of a request in token-level routing algorithms.
A request first enters an entry model and performs a decoding step.
After the forward pass, an algorithm-specific router decides whether the request should continue locally or be delegated to another model.
If the request is delegated, the current model gathers the necessary context, such as the latest generated tokens, and sends it to the target model.
The target model receives the request, continues decoding, and repeats the same route-send-receive process.
This lifecycle can be written as:
\[
\textbf{receive}
\rightarrow
\text{decode}
\rightarrow
\textbf{route}
\rightarrow
\textbf{send}
\rightarrow
\text{peer }\textbf{receive}
\rightarrow
\text{peer decode}
\rightarrow
\text{peer }\textbf{route}
\rightarrow
\cdots .
\]

Based on this observation, \name lets developers implement token-level routing algorithms by filling in only three components: \textbf{route}, \textbf{send}, and \textbf{receive}.
Developers program the algorithm from a simple request-centric view, while the \name runtime executes it with model-centric subservers.
This \emph{request-centric programming, model-centric execution} abstraction is the key principle behind \name's programming interface.

\subsection{Core Interface}
\label{sec:programming/interface}

We now detail the three components that developers implement in \name.

\textbf{route}\,(\texttt{result}) is called after each decoding step of an LLM.
It determines a destination index for each request in the batch based on the forward-pass result such as hidden states, next-token logits, and sampled tokens.
A destination of $0$ means that the request continues decoding locally, while a nonzero destination points to the corresponding peer model.

\textbf{send}\,(\texttt{req}) is called when \textbf{route} delegates a request.
It returns a customized \texttt{PeerReq} message that carries the request ID, the suffix of tokens that the peer model has not yet seen, a status field indicating whether the request is still active or has terminated, and any algorithm-specific fields needed by the receiver.
A position anchor \texttt{req.loc} is also attached to indicate the decoding position before the \texttt{PeerReq} is sent.
The receiver can use this anchor to determine the minimal suffix that should be sent back.
The exact token suffix is algorithm-dependent.
For example, a scheduler may exclude or include the just-sampled token depending on whether the peer should regenerate it.

\textbf{receive}\,(\texttt{peer\_req}) is called when a \texttt{PeerReq} arrives from a peer model.
It converts the \texttt{PeerReq} to the local request format so that the receiver model can continue decoding.
By default, \textbf{receive} commits all transferred tokens from the \texttt{PeerReq}, which is sufficient for most token-level routing algorithms.
Developers only need to override \textbf{receive} when the handoff semantics are algorithm-specific.
For example, an algorithm may attach additional state before resuming decoding.

\subsection{Running Examples}
\label{sec:programming/example}
\begin{figure}[t]
\centering
\begin{minipage}{0.95\linewidth}
\begin{lstlisting}[style=tokenrouter]
class RStitchSLM(BaseRouter):
    # self.tau (threshold) and self.peer_idx (LLM subserver index) come from the config.

    def route(self, result):
        # Delegate high-entropy tokens to the LLM peer; keep the rest local.
        probs = result.logits_output.next_token_logits.softmax(dim=-1)
        entropy = -(probs * probs.log()).sum(-1) / math.log(probs.shape[-1])
        destination = torch.zeros_like(entropy, dtype=torch.long)
        destination[entropy > self.tau] = self.peer_idx
        return destination.cpu()

    def send(self, req) -> PeerReq:
        # Drop the just-decoded uncertain token; ask the LLM to regenerate.
        token_ids = req.origin_input_ids + req.output_ids[:-1]
        peer_anchor = req.peer_locs[self.peer_idx] or 0
        return PeerReq(rid=req.rid, token_ids=token_ids[peer_anchor:],
                       loc=len(token_ids), finished=req.finished(), ...)

    # receive falls back to the default and accepts all returned tokens.
\end{lstlisting}
\end{minipage}
\caption{
Small-model side of the R-Stitch~\citep{chen2025rstitch} in \name.
\textbf{route} sends high-entropy tokens to the large model and keeps low-entropy tokens local.
\textbf{send} excludes the uncertain token and transfers only the peer-unseen suffix, letting the large model regenerate from that position.
\textbf{receive} uses the default accept-all conversion for returned large-model tokens.
The large-model side is identical, except that \texttt{>} is replaced by \texttt{<=} in \textbf{route} to dispatch low-entropy tokens to the SLM.
}
\label{fig:rstitch}
\end{figure}

Figure~\ref{fig:rstitch} shows an example of the interface for the small-model side of \emph{R-Stitch}~\citep{chen2025rstitch}.
The SLM delegates high-entropy tokens to the LLM, while the LLM sends low-entropy tokens back.

% Table~\ref{tab:methods} summarizes the \textbf{route} behavior of the five token-level routing algorithms evaluated in this work. Despite using different routing signals, each is expressible with compact \textbf{route} and \textbf{send} functions, and the default \textbf{receive} suffices for all of them. To minimize large-model usage, most algorithms invoke the LLM only for one token when the SLM requests help, so their LLM-side router always delegates control back after decoding; R-Stitch is the exception, with a symmetric design where the LLM returns control to the SLM only when its output entropy becomes low.
Table~\ref{tab:methods} summarizes the routing behavior of the five token-level routing algorithms evaluated in this work. 
Despite using different routing signals, each is expressible with \textbf{route} function. For CITER, R2R, and Co-LLM, Model $i$ invokes Model $j$ upon detecting an uncertain token, while Model $j$ returns control after generating one token. R-Stitch instead adopts a symmetric design, where the two models switch control based on token entropy. ME further generalizes token-level routing to more than two models, selecting the next model after every token according to the ensemble weights.

% \begin{table}[t]
% \centering
% \caption{
% Behavior of the four evaluated token-level routing algorithms.
% \textbf{route} shows which tokens are delegated to the peer model.
% \textbf{send} shows the transferred suffix: \textit{drop last} ignores the just-decoded uncertain token and asks the peer to regenerate, while \textit{keep last} forwards it as committed.
% \textbf{receive} defaults to accepting all peer tokens on both sides for all algorithms.
% }
% \label{tab:methods}
% \setlength{\tabcolsep}{3pt}
% \renewcommand{\arraystretch}{1.2}
% \begin{tabular}{@{}l l l c c c@{}}
% \toprule
% & \multicolumn{2}{c}{\textbf{route}: delegate \dots} & \multicolumn{2}{c}{\textbf{send}} & \textbf{receive} \\
% \cmidrule(lr){2-3} \cmidrule(lr){4-5} \cmidrule(lr){6-6}
% \textbf{Algorithm} & \textbf{SLM side} & \textbf{LLM side} & \textbf{SLM side} & \textbf{LLM side} & \textbf{Both sides} \\
% \midrule
% CITER~\citep{zheng2025citer}     & low-confidence tokens      & every token        & drop last & keep last & default \\
% R2R~\citep{fu2025r2r}            & predicted-divergent tokens & every token        & drop last & keep last & default \\
% R-Stitch~\citep{chen2025rstitch} & high-entropy tokens        & low-entropy tokens & drop last & keep last & default \\
% Co-LLM~\citep{shen2024learning}  & high-deferral tokens       & every token        & drop last & keep last & default \\
% \bottomrule
% \end{tabular}
% \end{table}
\begin{table}[t]
\centering
\small
\caption{
Routing behavior of the evaluated token-level routing algorithms.
Each entry under Model $i$ or Model $j$ specifies the condition under
which that model routes generation to another model.
}
\label{tab:methods}
\setlength{\tabcolsep}{5pt}
\renewcommand{\arraystretch}{1.2}

\begin{tabular}{@{}l c c c@{}}
\toprule
\textbf{Algorithm}
& \textbf{\#Models}
& \textbf{Model $i$}
& \textbf{Model $j$} \\
\midrule

CITER~\citep{zheng2025citer}
& 2
& low-confidence token
& after one token \\

R2R~\citep{fu2025r2r}
& 2
& predicted-divergent token
& after one token \\

R-Stitch~\citep{chen2025rstitch}
& 2
& high-entropy token
& low-entropy token \\

Co-LLM~\citep{shen2024learning}
& 2
& high-deferral token
& after one token \\

ME~\citep{fu2026rethinking}
& $\geq$2
& \multicolumn{2}{c}{after every token, according to ensemble weights} \\

\bottomrule
\end{tabular}
\end{table}

\section{System Design}
\label{sec:architecture}

This section presents \name's model-centric runtime, which aims to efficiently serve token-routed inference.
% While Section~\ref{sec:programming} exposes token-level routing algorithms through a request-centric programming interface, this section describes how \name executes them efficiently through a model-centric runtime.
Section~\ref{sec:architecture/overview} first gives an overview of the system architecture.
Section~\ref{sec:architecture/async} addresses \emph{step desynchronization} through asynchronous tri-loop execution and a handoff-resume mechanism.
Section~\ref{sec:architecture/delayed} addresses \emph{batch admission delay} with delayed batching, which trades a small waiting time for larger same-model batches.
Our implementation builds on SGLang~\citep{zheng2024sglang}; Appendix~\ref{appendix:opt} and Appendix~\ref{appendix:impl} describe more routing-specific engineering optimizations.

\subsection{System Overview}
\label{sec:architecture/overview}

\name's runtime is organized as multiple subservers behind a single external interface, as shown in Figure~\ref{fig:architecture}.
Internally, each subserver hosts one candidate LLM.
It owns a scheduler, an LLM runner with a private KV-cache pool, and three user-defined functions to construct the router, sender, and receiver.
Subservers progress independently and communicate through peer requests.
Their identical internal layout makes it easy to scale to more LLMs by launching additional subservers.

Externally, \name exposes a single server interface, which receives client requests and forwards them to a designated entry subserver.
% For simplicity, we refer to this entry point as Subserver~1.
The client-side protocol is therefore identical to that of standard single-LLM servers such as vLLM~\citep{kwon2023vllm} and SGLang~\citep{zheng2024sglang}, regardless of how many LLMs participate in routing behind the interface.
This design allows \name to serve as a drop-in replacement for existing single-LLM servers.

\subsection{Efficient Serving with Asynchronous Execution}
\label{sec:architecture/async}

\begin{figure}[tb]
    \centering
    \includegraphics[width=\linewidth]{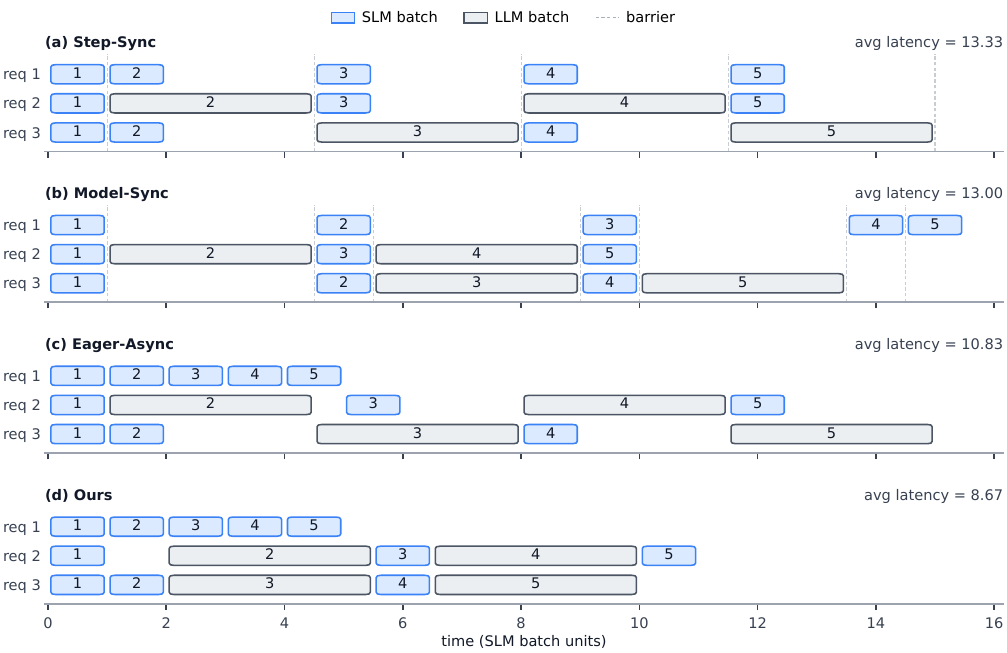}
    \caption{Scheduling strategies of token-level routing.
    (a) Step-Sync sets step barriers; the next step starts only after all models finish the current one.
    (b) Model-Sync runs one model at a time; the inactive model's requests stall.
    (c) Eager-Async processes peer requests immediately, causing large batch admission delay.
    (d) \name preserves asynchronous scheduling but delays LLM requests until enough same-model requests accumulate, amortizing the cost across a wider batch.}
    \label{fig:schedule}
\end{figure}

\name addresses the step desynchronization challenge by adding a decoupled inter-model loop and a handoff-resume mechanism on top of the model-centric architecture.
Figure~\ref{fig:schedule}(a) illustrates the desynchronization problem that arises when token-level routing directly follows the synchronous design of single-LLM serving systems. If requests at the same logical decoding step are synchronized, the whole batch must wait for the slowest model.

Modern serving systems also support speculative decoding, but its two models run in a tightly coupled, periodic pattern, whereas token-level routing invokes random model access.
Forcing routing algorithms to mimic speculative decoding serving would also incur severe idle time (Figure~\ref{fig:schedule}(b)).
To avoid global synchronization, \name adopts an asynchronous design (Figure~\ref{fig:schedule}(c, d)).

\xhdr{Decoupled tri-loop execution}
To support inter-model handoff, \name extends the standard execution structure of single-LLM serving with an additional decoupled loop.
% In a standard server, request admission and response run asynchronously with model execution.
% As shown in Figure~\ref{fig:architecture}, a \emph{client-server loop} (purple) receives requests and responses, while a \emph{decoding loop} (blue) schedules active requests and runs model forward passes.
As shown in Figure~\ref{fig:architecture}, a standard server has two asynchronous loops: a \emph{client-server loop} (purple) for request admission and response streaming, and a \emph{decoding loop} (blue) for request scheduling and model execution.
\name adds a third decoupled loop, the \emph{inter-model loop} (black), to send and receive peer requests between subservers.
This third loop allows a routed request to leave the local  batch and resume when the peer request returns.
While the routed request is being processed by the peer model, the remaining local requests continue executing at the pace of the local decoding loop.
When the peer request returns, the scheduler admits the resumed request back into the local batch.
Thus, each subserver progresses at its own pace, while routed requests move asynchronously across models.

\xhdr{Handoff and resume across models}
\name designs a handoff and resume mechanism for the decoupled inter-model loop. %, harnessing the fact that a routed request will return shortly.
The naive handoff treats a returning request as a fresh call, which redoes prefix matching and KV-cache allocation, incurring unnecessary costs.
To deal with this, \name introduces a \emph{pending} state, an intermediate status between the two standard states \emph{running} and \emph{finished}.
% Like \emph{finished}, a pending request is skipped by the scheduler; like \emph{running}, its serving state is fully preserved by parking the request in a side batch that retains its KV slot.
A \emph{pending} request is skipped by the scheduler, but its serving state is preserved. %by parking the request in a side batch that retains its KV slot.
At handoff, the routed request is marked as \emph{pending}.
Upon resumption, \emph{pending} is toggled to \emph{running} or \emph{finished} according to the peer's output.
If not finished, \name appends new tokens to the request and admits it into the local batch as if it had never left.
By skipping prefix matching and KV reallocation, handoff-resume reduces inter-model transitions to little more than a token append.

% \vspace{-15pt}
\subsection{Efficient Scheduling with Delayed Batching}
\label{sec:architecture/delayed}

While asynchronous execution removes synchronization across subservers, each subserver still needs a scheduling policy that handles sparse and irregular routed-token arrivals.
We define \emph{batch admission delay} as the waiting time between a request's arrival at a subserver and the start of its execution.
Figure~\ref{fig:schedule}(c) illustrates the issue under eager asynchronous scheduling.
If a subserver starts a new decoding step as soon as any peer request arrives, requests that arrive while the model is already processing a batch cannot be admitted until that step finishes.
For example, request~3's token~3 cannot start until the LLM finishes decoding request~2's token~2, incurring the admission delay from time~2 to time~4.5.
Because token-level routing causes frequent model switches and high per-step latency variance, such admission stalls occur repeatedly and lead to fragmented batching.

\xhdr{Delayed-batching scheduler}
To reduce admission delay, we propose a delayed-batching scheduler.
The key idea is to wait for a short, controlled interval so that more requests targeting the same model can be executed together.
Specifically, the scheduler buffers received requests and launches a batch only when the buffer size reaches a threshold $B$.
Figure~\ref{fig:schedule}(d) shows an example where the LLM subserver uses $B=2$, while the SLM subserver uses $B=1$.
Although an early-arriving request waits longer than under eager-async in Figure~\ref{fig:schedule}(c), later-arriving requests can be admitted with much less waiting, so the workload's average batch admission delay drops. 

\xhdr{Throughput-optimal threshold}
\name analytically determines the throughput-optimal delayed-batching threshold $B^*$.
This requires balancing a strict trade-off: too small $B$ causes large batch admission delay, while too large $B$ traps requests in the queue and starves the SLM of active work. 
We model the routing process as a Discrete-Time Markov Chain (DTMC).
Given a token-level routing algorithm, concurrency $N$, routing probability $\mathbf{P}$, and per-step decoding latency $L_i$ of each LLM, 
our model derives the throughput as a function of $B$ and searches the threshold $B^*$ that maximizes the throughput. 
Full details are provided in Appendix~\ref{appendix:math_model}.

\section{Experiment}
\label{sec:experiment}

\subsection{Setup}
\label{sec:exp/setup}

\xhdr{Token-level routing algorithms}
\label{sec:exp:setup:alg}
We evaluate five token-level routing algorithms in \name, namely CITER~\citep{zheng2025citer}, R2R~\citep{fu2025r2r}, R-Stitch~\citep{chen2025rstitch}, Co-LLM~\citep{shen2024learning}, and ME~\citep{fu2026rethinking}.
CITER, R2R, R-Stitch, and Co-LLM route between Qwen3-0.6B and Qwen3-32B~\citep{yang2025qwen3}, while ME routes among Qwen3-0.6B, Qwen3-8B, and Qwen3-32B.
For CITER, we train the router and set $\tau=0.98$.
For R2R, we use the official router and threshold.
For R-Stitch, we set $\tau=0.03$.
For Co-LLM, we fine-tune Qwen3-0.6B and set $\eta=0.15$.
For ME, we set the ensemble weight to 0.5, 0.3, and 0.2.
We also evaluate each algorithm using the model pair and the dataset from its original paper, as summarized in Table~\ref{tab:model_pair_benchmark}.

\xhdr{Hardware}
All experiments run on an 8$\times$A100-80G GPU server.
For two-model algorithms, the small model is deployed on one GPU and the large model on two GPUs with tensor parallelism, with two models sharing two GPUs via CUDA MPS.
For ME, each model is deployed on one GPU.

\xhdr{Baselines}
We compare \name against two baselines.
\emph{Official Code} refers to the algorithm's released implementation, available for all but R-Stitch (whose code is not released). Among these, Co-LLM uses a per-model vLLM~\citep{kwon2023vllm} server with continuous batching, while CITER and R2R are inference engines without continuous batching.
\emph{Std.\ Serving} is a standard SGLang-based baseline that we build for a fair comparison. It uses one SGLang server per LLM; each server returns one token at a time, and an external dispatcher routes each token between servers.
% For the LLM-only setting (no routing), \emph{Official Code} refers to a HuggingFace Transformers implementation, and \emph{Std.\ Serving} to a single SGLang server running the large model with a tensor parallelism degree of 2.

\xhdr{Workloads}
We evaluate three workloads spanning different input/output lengths.
(1) For \emph{Low-effort reasoning} (short input, short output), we use AIME2024~\citep{aime} prompts ($\sim$100 tokens) with a maximum output length of 2{,}048.
(2) For \emph{High-effort reasoning} (short input, long output), we use AIME2024 prompts with a maximum output length to 8{,}192, keeping only problems whose Qwen3-32B solutions exceed 8{,}192 tokens.
(3) For \emph{Agentic} tasks (multi-turn), we use SWE-Smith~\citep{yang2025swe} trajectories of around 8{,}192 input tokens with a maximum output length of 1{,}024.

\xhdr{Metrics}
% We report throughput, measured in output tokens per second, as the batch size $N$ varies.
% For inference engines, we form a batch of $N$ requests and compute throughput as the total generated tokens divided by batch processing time.
% For serving systems, we launch $N$ workers; each worker issues a new request as soon as the previous one returns, maintaining $N$ in-flight requests at the server.
% Throughput is then the sum of per-worker token generation rates.
We report the throughput and the end-to-end request completion latency.
For inference engines, we form a batch of $N$ requests and process the batch.
For serving systems, we launch $N$ workers, each of which issues a new request as soon as the previous one completes.
We measure throughput as the total number of generated output tokens divided by elapsed time.

\subsection{Serving Efficiency}

\begin{figure}[t]
    \centering
    \includegraphics[width=1\linewidth]{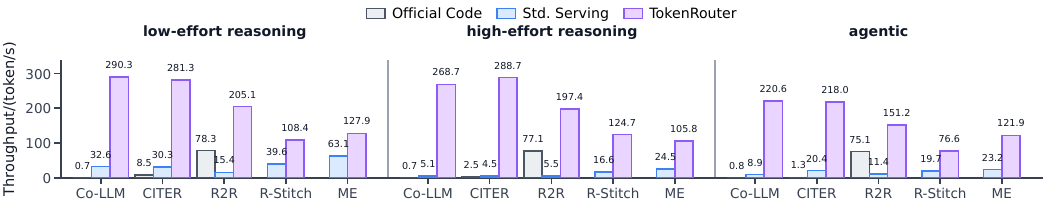}
    \caption{Throughput (token/s) comparison across serving systems at concurrency 4.}
    \label{fig:throughput_bs4}
\end{figure}

\begin{figure}[t]
    \centering
    \includegraphics[width=1\linewidth]{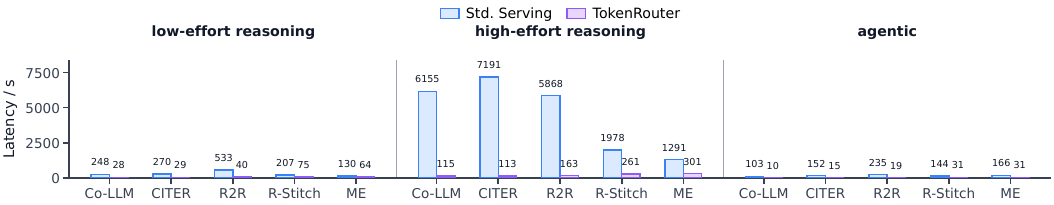}
    \caption{Latency (s) comparison across serving systems at concurrency 4.}
    \label{fig:latency_bs4}
\end{figure}

\begin{table*}[t]
\caption{TTFT, throughput, and end-to-end latency at concurrency 4, comparing across serving systems under original settings.}
\label{tab:e2e_latency_bs4}
\centering
\begin{tabular}{ll|ccc}
\toprule
\textbf{Algorithm} & \textbf{Implementation}
& \textbf{Throughput/(token/s)} & \textbf{TTFT/s} & \textbf{Latency/s} \\
\midrule
\multirow{3}{*}{R2R}
& LLM-only
& 145.30
& 0.13
& 411.98 \\

& Official Code
& 89.62
& 0.11
& 751.15 \\

& TokenRouter
& 244.56
& 0.11
& \textbf{270.19} \\

\midrule

\multirow{3}{*}{CITER}
& LLM-only
& 123.72
& 0.083
& 2.68 \\

& Official Code
& 17.16
& 0.036
& 7.64 \\

& TokenRouter
& 149.31
& 0.036
& \textbf{0.48} \\

\midrule

\multirow{3}{*}{Co-LLM}
& LLM-only
& 134.79
& 0.069
& 13.39 \\

& Official Code
& 3.46 
& 1.14
& 247.61 \\

& TokenRouter
& 76.02
& 0.067
& \textbf{11.46} \\

\midrule

\multirow{3}{*}{R-Stitch}
& LLM-only
& 150.41
& 0.15
& 360.76 \\

& Official Code
& N/A & N/A & N/A \\

& TokenRouter
& 140.58
& 0.067
& \textbf{151.30} \\

\bottomrule

\end{tabular}

\end{table*}

As shown in Figure~\ref{fig:throughput_bs4} and~\ref{fig:latency_bs4}, across all 15 algorithm--workload combinations, \name consistently outperforms the baselines, improving throughput by 2.01--64.15$\times$ compared with the stronger baseline.
\name also reduces the end-to-end latency of Std. Serving by 2.03--63.64$\times$.
Although Std. Serving substantially improves upon the official Co-LLM and CITER implementations through asynchronous execution, it lacks \name's handoff-and-resume mechanism.
Consequently, the serving state of a routed request cannot be preserved across model switches, and each returning token must be treated as a new request.
This incurs substantial overhead for operations such as prefix matching and KV-cache updates.
We further break down the per-step latency of Std. Serving to quantify this overhead in Table~\ref{tab:slm_latency_breakdown}.
% This advantage is most pronounced for CITER under high-effort reasoning, where \name reaches 288.67 token/s versus 4.49 token/s with standard serving, a 64.29$\times$ improvement, while reducing latency from 7{,}190.60 seconds to 112.87 seconds.

\name also scales gracefully with output length.
When we compare low-effort and high-effort reasoning workloads (max output 2{,}048 vs.\ 8{,}192 tokens), \name's throughput holds steady while Std. Serving loses 58.1--85.2\% of its throughput.

% \begin{wrapfigure}{r}{0.48\linewidth}
%     \centering
%     \vspace{-15pt}
%     \includegraphics[width=\linewidth]{figure/experiment/tradeoff.pdf}
%     \caption{Throughput--interactivity trade-off of \name and the official R2R implementation under different batch sizes. Each marker represents one batch size.}
%     \label{fig:throughput_interactivity}
%     \vspace{-12pt}
% \end{wrapfigure}

When evaluated on the original settings of the token-level routing algorithms, as shown in Table~\ref{tab:e2e_latency_bs4},  \name reduces latency by 2.78--21.61$\times$ over the official implementations while substantially improving throughput by 2.73--21.97$\times$.
These results show that \name efficiently supports diverse token-level routing algorithms under their original algorithmic settings.

\textbf{Throughput--Speed Trade-off. }
Figure~\ref{fig:throughput_interactivity} evaluates the trade-off between throughput and per-user speed as the concurrency increases from 1 to 16.
When the concurrency increases from 1 to 16, \name's throughput increases by 8.61$\times$, while retaining 51.7\% of its single-user speed.
In contrast, official R2R reaches only 5.14$\times$ and retains 31.3\% of its single-user speed.
Notably, at concurrency 16, \name achieves 18.58$\times$ higher throughput than official R2R at concurrency 1, while still providing 1.13$\times$ higher per-user speed.
This means that, \name delivers substantially higher throughput even under a stricter per-user speed Service Level Objective (SLO).

\subsection{Ablation Study}
We briefly summarize the conclusions of the ablation study here, with details in Appendix~\ref{sec:appendix/ablation}.

% \begin{figure}[t]
%     \centering
%     \includegraphics[width=0.48\linewidth]{figure/experiment/throughput_accuracy_tradeoff.pdf}
%     \caption{Throughput--accuracy trade-off by varying thresholds while keeping batch sizes fixed. QR denotes query-level routing from RouteLLM.}
%     \label{fig:experiment_analysis}
% \end{figure}
\begin{figure*}[t]
    \centering

    % ===== Left: Table =====
    \begin{minipage}[t]{0.49\textwidth}
        \centering
        \small

        \vspace{-73pt}

        \resizebox{6.9cm}{!}{
        \begin{tabular}{llccc}
        \toprule
        \textbf{SLM} & \textbf{LLM} & \textbf{R2R} &
        \textbf{\name} & \textbf{Gain} \\
        \midrule
        0.6B & 8B  & 105.1  & 336.98 & 3.21$\times$ \\
        0.6B & 32B & 77.08  & 197.40 & 2.56$\times$ \\
        1.7B & 8B  & 103.37 & 285.41 & 2.76$\times$ \\
        4B   & 8B  & 106.67 & 211.81 & 1.99$\times$ \\
        \bottomrule
        \end{tabular}
        }
        \vspace{4.5pt}
        \captionof{table}{R2R vs.\ \name across SLM-LLM pairs.
        Throughput is reported in token/s.}
        \label{tab:r2r_tokenrouter_gain_app}
    \end{minipage}
    \hfill
    % ===== Right: Figure =====
    \begin{minipage}[t]{0.49\textwidth}
        \centering
        
        \includegraphics[width=6.6cm]{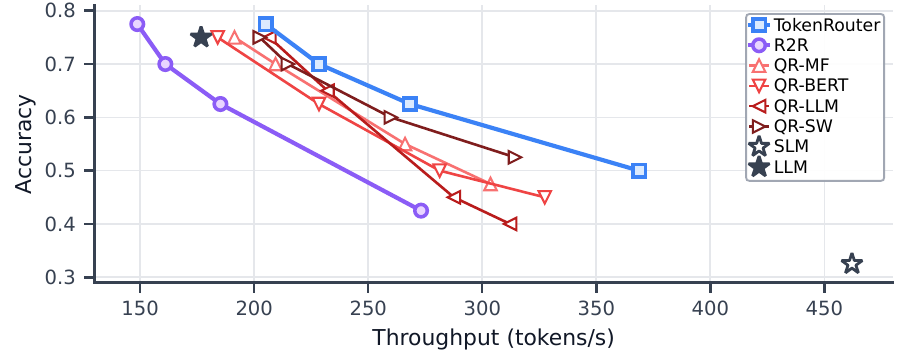}
        \vspace{-8pt}
        \captionof{figure}{Throughput--accuracy trade-off.
        QR denotes query-level routing from RouteLLM.}
        \label{fig:experiment_analysis}
    \end{minipage}

\end{figure*}

\xhdr{Gain breakdown}
Figure~\ref{fig:gain_breakdown} breaks down \name's throughput at concurrency 8.
Engineering optimizations with extended CUDA graphs raise throughput from the base implementation's 132.78 to 230.79 token/s, or 1.71$\times$ the official R2R implementation (134.9 token/s).
Asynchronous execution and delayed batching further increase throughput to 296.86 and 372.48 token/s, respectively, reaching 2.76$\times$ overall.
Table~\ref{tab:gain_source} provides additional ablations at concurrency 1 and 4.

% \textbf{Mathematical modeling accuracy.} 
% Figure~\ref{fig:experiment_analysis}(a) plots measured throughput as a function of the delayed-batching threshold $B^{*}$, alongside predictions from our closed-form mathematical model with a routing probability of 0.2.
% The model closely tracks the empirical curve, and its predicted optimum matches the measured peak, providing accurate guidance for choosing $B^{*}$ without per-workload tuning.
% The model further shows that the optimal $B^{*}$ increases with the batch size $N$, the LLM-to-SLM request ratio, and the LLM-to-SLM latency ratio (Appendix~\ref{appendix:execution_threshold}).

\xhdr{Effect of model pairs}
% Table~\ref{tab:r2r_tokenrouter_gain_app} compares \name with R2R across four model pairs: 0.6B-8B, 0.6B-32B, 1.7B-8B, and 4B-8B. Across these pairs, \name consistently improves throughput over R2R, with gains ranging from 1.99$\times$ to 3.21$\times$ on the high-effort reasoning workload.
Table~\ref{tab:r2r_tokenrouter_gain_app} evaluates \name against the most competitive R2R system baseline across diverse model pairs.
Compared with the official implementation of R2R, \name achieves 1.99--3.21$\times$ throughput improvement across different model pairs. These results suggest that \name's optimizations are effective across different SLM and LLM scales.

\xhdr{Parallel strategy}
Table~\ref{tab:parallelism} shows that increasing LLM tensor parallelism size is vital for throughput, while the SLM tensor parallelism size has little effect. Thus, unless otherwise specified, we set LLM tensor parallelism size to 2 and SLM tensor parallelism size to 1.

% \begin{figure}[t]
%     \centering
%     \begin{subfigure}[t]{0.65\textwidth}
%         \centering
%         \includegraphics[width=\linewidth]{figure/experiment/delayed_batching.pdf}
%         \label{fig:delayed_batching}
%     \end{subfigure}
%     \hfill
%     \begin{subfigure}[t]{0.32\textwidth}
%         \centering
%         \includegraphics[width=\linewidth]{figure/experiment/throughput_accuracy_tradeoff.pdf}
%         \label{fig:throughput_accuracy_tradeoff}
%     \end{subfigure}
%     \vspace{-12pt}
%     \caption{(a) Throughput as a function of the delayed-batching threshold $B^{*}$, showing an interior optimum (analyzed in Appendix~\ref{appendix:non-overlap-model}). 
%     (b) \name advances the Pareto frontier for the throughput-accuracy trade-off in token-level routing. QR denotes query routing from RouteLLM.}
%     \label{fig:experiment_analysis}
% \end{figure}

% \begin{wrapfigure}{r}{0.48\linewidth}
%     \centering
%     \vspace{-15pt}
%     \includegraphics[width=\linewidth]{figure/experiment/throughput_accuracy_tradeoff.pdf}
%     \caption{
%     \name advances the Pareto frontier for the throughput-accuracy trade-off in token-level routing. QR denotes query routing from RouteLLM.}
%     \label{fig:experiment_analysis}
%     \vspace{-12pt}
% \end{wrapfigure}

\subsection{Advancing the Pareto Frontier}
We evaluate the throughput-accuracy trade-off of \name, the official R2R implementation, and query-level routing baselines on AMC23, using the Qwen3-0.6B and Qwen3-32B pair with greedy decoding and a maximum output length of 8{,}192 tokens. All requests are sent concurrently; throughput is computed per request as output tokens divided by elapsed time; and SLM/LLM memory fractions are held equal across systems for a fair comparison. Figure~\ref{fig:experiment_analysis} shows that \name shifts token-level routing to a new Pareto frontier, making fine-grained routing more competitive than coarse-grained routing on state-of-the-art frameworks such as SGLang.

\section{Conclusion}
\label{sec:conclusion}

We presented \name, an efficient serving system for token-level LLM routing that decouples routing algorithms from serving-system optimization.
\name lets developers express routing semantics through a request-centric route-send-receive interface, while the runtime handles asynchronous model-centric execution.
We overcome the step desynchronization and batch admission delay challenges for efficient serving with asynchronous execution and the delayed-batching scheduler.
Across routing algorithms, workloads, and model pairs, \name consistently achieves 2.01--64.15$\times$ higher decoding throughput than existing implementations, providing a strong foundation for future token-routing research.

\xhdr{Limitation and future work} Our mathematical model assumes that the number of tokens between two consecutive sends follows an geometric distribution, which holds for most token-level routing algorithms. 
Handling corner cases that violate this assumption is left for future work.

% Acknowledgments
\clearpage

\begin{ack}
This work was supported by National Natural Science Foundation of China (No. 62325405, 62104128, U19B2019, U21B2031, 61832007, 62204164, 92364201), Tsinghua EE Xilinx AI Research Fund, and Beijing National Research Center for Information Science and Technology (BNRist).
We thank for all the support from Infinigence-AI.
\end{ack}

% Reference
\medskip

{
\small
\bibliographystyle{unsrtnat}
\bibliography{refs}

@article{zheng2025citer,
  title={Citer: Collaborative inference for efficient large language model decoding with token-level routing},
  author={Zheng, Wenhao and Chen, Yixiao and Zhang, Weitong and Kundu, Souvik and Li, Yun and Liu, Zhengzhong and Xing, Eric P and Wang, Hongyi and Yao, Huaxiu},
  journal={arXiv preprint arXiv:2502.01976},
  year={2025}
}

@article{jin2024collaborative,
  title={Collaborative decoding of critical tokens for boosting factuality of large language models},
  author={Jin, Lifeng and Peng, Baolin and Song, Linfeng and Mi, Haitao and Tian, Ye and Yu, Dong},
  journal={arXiv preprint arXiv:2402.17982},
  year={2024}
}

@inproceedings{shen2024learning,
  title={Learning to decode collaboratively with multiple language models},
  author={Shen, Zejiang and Lang, Hunter and Wang, Bailin and Kim, Yoon and Sontag, David},
  booktitle={Proceedings of the 62nd Annual Meeting of the Association for Computational Linguistics (Volume 1: Long Papers)},
  pages={12974--12990},
  year={2024}
}

@article{fu2025r2r,
  title={R2r: Efficiently navigating divergent reasoning paths with small-large model token routing},
  author={Fu, Tianyu and Ge, Yi and You, Yichen and Liu, Enshu and Yuan, Zhihang and Dai, Guohao and Yan, Shengen and Yang, Huazhong and Wang, Yu},
  journal={arXiv preprint arXiv:2505.21600},
  year={2025}
}

@article{huang2026relayllm,
  title={Relayllm: Efficient reasoning via collaborative decoding},
  author={Huang, Chengsong and Zheng, Tong and Huang, Langlin and Li, Jinyuan and Liu, Haolin and Huang, Jiaxin},
  journal={arXiv preprint arXiv:2601.05167},
  year={2026}
}

@article{xiong2026fusionroute,
  title={Token-Level LLM Collaboration via FusionRoute},
  author={Xiong, Nuoya and Zhou, Yuhang and Zeng, Hanqing and Chen, Zhaorun and Huang, Furong and Bi, Shuchao and Zhang, Lizhu and Zhao, Zhuokai},
  journal={arXiv preprint arXiv:2601.05106},
  year={2026}
}

@article{rai2026taro,
  title={TARo: Token-level Adaptive Routing for LLM Test-time Alignment},
  author={Rai, Arushi and Zhang, Qiang and Zeng, Hanqing and Zhang, Yunkai and Tamboli, Dipesh and Fan, Xiangjun and Zhao, Zhuokai},
  journal={arXiv preprint arXiv:2603.18411},
  year={2026}
}

@article{chen2025rstitch,
  title={R-stitch: Dynamic trajectory stitching for efficient reasoning},
  author={Chen, Zhuokun and Chen, Zeren and He, Jiahao and Sheng, Lu and Tan, Mingkui and Cai, Jianfei and Zhuang, Bohan},
  journal={arXiv preprint arXiv:2507.17307},
  year={2025}
}

@article{zeng2026glimprouter,
  title={Glimprouter: Efficient collaborative inference by glimpsing one token of thoughts},
  author={Zeng, Wenhao and Zhang, Xuteng and Shi, Yuling and Hu, Chao and Chen, Yuting and Shen, Beijun and Gu, Xiaodong},
  journal={arXiv preprint arXiv:2601.05110},
  year={2026}
}

@article{akhauri2025splitreason,
  title={Splitreason: Learning to offload reasoning},
  author={Akhauri, Yash and Fei, Anthony and Chang, Chi-Chih and AbouElhamayed, Ahmed F and Li, Yueying and Abdelfattah, Mohamed S},
  journal={arXiv preprint arXiv:2504.16379},
  year={2025}
}

@article{singh2025gpt5,
  title={Openai gpt-5 system card},
  author={Singh, Aaditya and Fry, Adam and Perelman, Adam and Tart, Adam and Ganesh, Adi and El-Kishky, Ahmed and McLaughlin, Aidan and Low, Aiden and Ostrow, AJ and Ananthram, Akhila and others},
  journal={arXiv preprint arXiv:2601.03267},
  year={2025}
}

@article{zheng2024sglang,
  title={Sglang: Efficient execution of structured language model programs},
  author={Zheng, Lianmin and Yin, Liangsheng and Xie, Zhiqiang and Sun, Chuyue and Huang, Jeff and Yu, Cody H and Cao, Shiyi and Kozyrakis, Christos and Stoica, Ion and Gonzalez, Joseph E and others},
  journal={Advances in neural information processing systems},
  volume={37},
  pages={62557--62583},
  year={2024}
}

@inproceedings{kwon2023vllm,
  title={Efficient memory management for large language model serving with pagedattention},
  author={Kwon, Woosuk and Li, Zhuohan and Zhuang, Siyuan and Sheng, Ying and Zheng, Lianmin and Yu, Cody Hao and Gonzalez, Joseph and Zhang, Hao and Stoica, Ion},
  booktitle={Proceedings of the 29th symposium on operating systems principles},
  pages={611--626},
  year={2023}
}

@misc{su2025toolorchestraelevatingintelligenceefficient,
      title={ToolOrchestra: Elevating Intelligence via Efficient Model and Tool Orchestration},
      author={Hongjin Su and Shizhe Diao and Ximing Lu and Mingjie Liu and Jiacheng Xu and Xin Dong and Yonggan Fu and Peter Belcak and Hanrong Ye and Hongxu Yin and Yi Dong and Evelina Bakhturina and Tao Yu and Yejin Choi and Jan Kautz and Pavlo Molchanov},
      year={2025},
      eprint={2511.21689},
      archivePrefix={arXiv},
      primaryClass={cs.CL},
      url={https://arxiv.org/abs/2511.21689},
}

@misc{chen2023frugalgptuselargelanguage,
      title={FrugalGPT: How to Use Large Language Models While Reducing Cost and Improving Performance},
      author={Lingjiao Chen and Matei Zaharia and James Zou},
      year={2023},
      eprint={2305.05176},
      archivePrefix={arXiv},
      primaryClass={cs.LG},
      url={https://arxiv.org/abs/2305.05176},
}

@misc{ding2024hybridllmcostefficientqualityaware,
      title={Hybrid LLM: Cost-Efficient and Quality-Aware Query Routing},
      author={Dujian Ding and Ankur Mallick and Chi Wang and Robert Sim and Subhabrata Mukherjee and Victor Ruhle and Laks V. S. Lakshmanan and Ahmed Hassan Awadallah},
      year={2024},
      eprint={2404.14618},
      archivePrefix={arXiv},
      primaryClass={cs.LG},
      url={https://arxiv.org/abs/2404.14618},
}

@inproceedings{NEURIPS2024_ecda225c,
 author = {Aggarwal, Pranjal and Madaan, Aman and Anand, Ankit and Potharaju, Srividya Pranavi and Mishra, Swaroop and Zhou, Pei and Gupta, Aditya and Rajagopal, Dheeraj and Kappaganthu, Karthik and Yang, Yiming and Upadhyay, Shyam and Faruqui, Manaal and Mausam, Mausam},
 booktitle = {Advances in Neural Information Processing Systems},
 doi = {10.52202/079017-4164},
 editor = {A. Globerson and L. Mackey and D. Belgrave and A. Fan and U. Paquet and J. Tomczak and C. Zhang},
 pages = {131000--131034},
 publisher = {Curran Associates, Inc.},
 title = {AutoMix: Automatically Mixing Language Models},
 url = {https://proceedings.neurips.cc/paper_files/paper/2024/file/ecda225cb187b40ea8edc1f46b03ffda-Paper-Conference.pdf},
 volume = {37},
 year = {2024}
}

@misc{zhao2024eagleefficienttrainingfreerouter,
      title={Eagle: Efficient Training-Free Router for Multi-LLM Inference},
      author={Zesen Zhao and Shuowei Jin and Z. Morley Mao},
      year={2024},
      eprint={2409.15518},
      archivePrefix={arXiv},
      primaryClass={cs.LG},
      url={https://arxiv.org/abs/2409.15518},
}

@misc{hu2024routerbenchbenchmarkmultillmrouting,
      title={RouterBench: A Benchmark for Multi-LLM Routing System},
      author={Qitian Jason Hu and Jacob Bieker and Xiuyu Li and Nan Jiang and Benjamin Keigwin and Gaurav Ranganath and Kurt Keutzer and Shriyash Kaustubh Upadhyay},
      year={2024},
      eprint={2403.12031},
      archivePrefix={arXiv},
      primaryClass={cs.LG},
      url={https://arxiv.org/abs/2403.12031},
}

@misc{huang2025routerevalcomprehensivebenchmarkrouting,
      title={RouterEval: A Comprehensive Benchmark for Routing LLMs to Explore Model-level Scaling Up in LLMs},
      author={Zhongzhan Huang and Guoming Ling and Yupei Lin and Yandong Chen and Shanshan Zhong and Hefeng Wu and Liang Lin},
      year={2025},
      eprint={2503.10657},
      archivePrefix={arXiv},
      primaryClass={cs.CL},
      url={https://arxiv.org/abs/2503.10657},
}

@misc{lu2023routingexpertefficientrewardguided,
      title={Routing to the Expert: Efficient Reward-guided Ensemble of Large Language Models},
      author={Keming Lu and Hongyi Yuan and Runji Lin and Junyang Lin and Zheng Yuan and Chang Zhou and Jingren Zhou},
      year={2023},
      eprint={2311.08692},
      archivePrefix={arXiv},
      primaryClass={cs.CL},
      url={https://arxiv.org/abs/2311.08692},
}

@misc{feng2025graphroutergraphbasedrouterllm,
      title={GraphRouter: A Graph-based Router for LLM Selections},
      author={Tao Feng and Yanzhen Shen and Jiaxuan You},
      year={2025},
      eprint={2410.03834},
      archivePrefix={arXiv},
      primaryClass={cs.AI},
      url={https://arxiv.org/abs/2410.03834},
}

@misc{wang2024mixtureofagentsenhanceslargelanguage,
      title={Mixture-of-Agents Enhances Large Language Model Capabilities},
      author={Junlin Wang and Jue Wang and Ben Athiwaratkun and Ce Zhang and James Zou},
      year={2024},
      eprint={2406.04692},
      archivePrefix={arXiv},
      primaryClass={cs.CL},
      url={https://arxiv.org/abs/2406.04692},
}

@article{ong2024routellm,
  title={Routellm: Learning to route llms with preference data},
  author={Ong, Isaac and Almahairi, Amjad and Wu, Vincent and Chiang, Wei-Lin and Wu, Tianhao and Gonzalez, Joseph E and Kadous, M Waleed and Stoica, Ion},
  journal={arXiv preprint arXiv:2406.18665},
  year={2024}
}

@article{yang2025qwen3,
  title={Qwen3 technical report},
  author={Yang, An and Li, Anfeng and Yang, Baosong and Zhang, Beichen and Hui, Binyuan and Zheng, Bo and Yu, Bowen and Gao, Chang and Huang, Chengen and Lv, Chenxu and others},
  journal={arXiv preprint arXiv:2505.09388},
  year={2025}
}

@misc{aime,
    title = {American Invitational Mathematics Examination},
    url = {https://artofproblemsolving.com/},
    author = {AIME},
    year = {2025}
}

@article{yang2025swe,
  title={Swe-smith: Scaling data for software engineering agents},
  author={Yang, John and Lieret, Kilian and Jimenez, Carlos E and Wettig, Alexander and Khandpur, Kabir and Zhang, Yanzhe and Hui, Binyuan and Press, Ofir and Schmidt, Ludwig and Yang, Diyi},
  journal={arXiv preprint arXiv:2504.21798},
  year={2025}
}

@inproceedings{yu2022orca,
  author = {Gyeong-In Yu and Joo Seong Jeong and Geon-Woo Kim and Soojeong Kim and Byung-Gon Chun},
  title = {Orca: A Distributed Serving System for {Transformer-Based} Generative Models},
  booktitle = {16th USENIX Symposium on Operating Systems Design and Implementation (OSDI 22)},
  year = {2022},
  isbn = {978-1-939133-28-1},
  address = {Carlsbad, CA},
  pages = {521--538},
  url = {https://www.usenix.org/conference/osdi22/presentation/yu},
  publisher = {USENIX Association},
  month = jul
}

@article{fu2025think,
  title={Think-at-Hard: Selective Latent Iterations to Improve Reasoning Language Models},
  author={Fu, Tianyu and You, Yichen and Chen, Zekai and Dai, Guohao and Yang, Huazhong and Wang, Yu},
  journal={arXiv preprint arXiv:2511.08577},
  year={2025}
}

@misc{leviathan2023fastinferencetransformersspeculative,
      title={Fast Inference from Transformers via Speculative Decoding}, 
      author={Yaniv Leviathan and Matan Kalman and Yossi Matias},
      year={2023},
      eprint={2211.17192},
      archivePrefix={arXiv},
      primaryClass={cs.LG},
      url={https://arxiv.org/abs/2211.17192}, 
}

@misc{cai2024medusasimplellminference,
      title={Medusa: Simple LLM Inference Acceleration Framework with Multiple Decoding Heads}, 
      author={Tianle Cai and Yuhong Li and Zhengyang Geng and Hongwu Peng and Jason D. Lee and Deming Chen and Tri Dao},
      year={2024},
      eprint={2401.10774},
      archivePrefix={arXiv},
      primaryClass={cs.LG},
      url={https://arxiv.org/abs/2401.10774}, 
}

@misc{li2025eaglespeculativesamplingrequires,
      title={EAGLE: Speculative Sampling Requires Rethinking Feature Uncertainty}, 
      author={Yuhui Li and Fangyun Wei and Chao Zhang and Hongyang Zhang},
      year={2025},
      eprint={2401.15077},
      archivePrefix={arXiv},
      primaryClass={cs.LG},
      url={https://arxiv.org/abs/2401.15077}, 
}

@inproceedings{liao2026alpharouter,
  title={AlphaRouter: Token-level Routing Between SLM and LLM with Reinforcement Learning and Tree Search},
  author={Liao, Siteng and Liang, Yuzhu and Rao, Hengzhong and Luo, Xizhao and Wang, Tian},
  booktitle={Forty-third International Conference on Machine Learning},
  year={2026}
}

@misc{cursor2026router,
  author       = {{Cursor Team}},
  title        = {Introducing Cursor Router},
  year         = {2026},
  month        = jul,
  day          = {22},
  howpublished = {\url{https://cursor.com/blog/router}},
  note         = {Accessed: 2026-07-25}
}

@article{fu2025cache,
  title={Cache-to-cache: Direct semantic communication between large language models},
  author={Fu, Tianyu and Min, Zihan and Zhang, Hanling and Yan, Jichao and Dai, Guohao and Ouyang, Wanli and Wang, Yu},
  journal={arXiv preprint arXiv:2510.03215},
  year={2025}
}

@article{fu2026rethinking,
  title={Rethinking LLM Ensembling from the Perspective of Mixture Models},
  author={Fu, Jiale and Jiang, Yuchu and Wu, Peijun and Liu, Chonghan and Zhou, Joey Tianyi and Yang, Xu},
  journal={arXiv preprint arXiv:2605.00419},
  year={2026}
}

@inproceedings{talmor2019commonsenseqa,
  title={Commonsenseqa: A question answering challenge targeting commonsense knowledge},
  author={Talmor, Alon and Herzig, Jonathan and Lourie, Nicholas and Berant, Jonathan},
  booktitle={Proceedings of the 2019 Conference of the North American Chapter of the Association for Computational Linguistics: Human Language Technologies, Volume 1 (Long and Short Papers)},
  pages={4149--4158},
  year={2019}
}

@article{cobbe2021training,
  title={Training verifiers to solve math word problems},
  author={Cobbe, Karl and Kosaraju, Vineet and Bavarian, Mohammad and Chen, Mark and Jun, Heewoo and Kaiser, Lukasz and Plappert, Matthias and Tworek, Jerry and Hilton, Jacob and Nakano, Reiichiro and others},
  journal={arXiv preprint arXiv:2110.14168},
  year={2021}
}
}

%%%%%%%%%%%%%%%%%%%%%%%%%%%%%%%%%%%%%%%%%%%%%%%%%%%%%%%%%%%%
\clearpage
\appendix
\begin{table}[t]
\centering
\caption{Model pairs and benchmarks used by different token-level routing algorithms.}
\label{tab:model_pair_benchmark}
\begin{tabular}{cll}

\toprule

\textbf{System} & \textbf{Model Pair} & \textbf{Benchmark} \\

\midrule
R2R
& DeepSeek-R1-Distill-Qwen-1.5B / 32B
& AIME~\citep{aime} \\
CITER
& Qwen2-1.5B / 72B
& CommonsenseQA~\citep{talmor2019commonsenseqa} \\
Co-LLM
& LLaMA2-7B (tuned) / 70B
& GSM8K~\citep{cobbe2021training} \\
R-Stitch
& L1-1.5B-Short / QwQ-32B
& AIME~\citep{aime} \\
\bottomrule
\end{tabular}
\end{table}

\begin{figure}[h]
    \centering
    \includegraphics[width=1\linewidth]{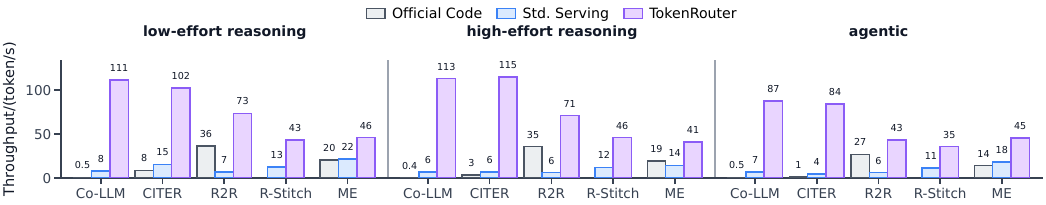}
    \caption{Throughput (token/s) comparison across serving systems at concurrency 1.}
    \label{fig:throughput_bs1}
\end{figure}

\begin{figure}[h]
    \centering
    \includegraphics[width=1\linewidth]{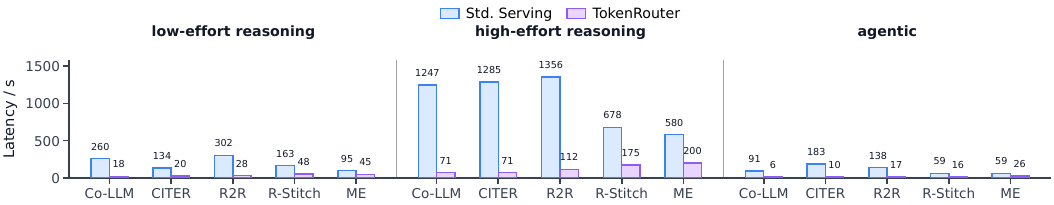}
    \caption{Latency (s) comparison across serving systems at concurrency 1.}
    \label{fig:latency_bs1}
\end{figure}

\section{Additional Experiment Results}
\label{appendix:exp_reuslt}

\subsection{Additional Concurrency}

Figure~\ref{fig:throughput_bs1} and~\ref{fig:latency_bs1} compare different implementations of five token-level routing algorithms across three workloads at concurrency 1.  \name consistently outperforms standard serving in all settings, achieving 2.12--19.74$\times$ higher throughput and a 2.12--18.83$\times$ latency speedup; it also provides 1.60--19.74$\times$ higher throughput than the stronger available baseline between the official code and standard serving.

As shown in Table~\ref{tab:e2e_latency_bs1}, under the original settings of each token-level routing algorithm at concurrency 1, \name consistently improves serving efficiency over the official implementations. It achieves 1.93--10.20$\times$ higher throughput and reduces end-to-end latency by 2.30--10.09$\times$ for R2R, CITER, and Co-LLM, while maintaining comparable TTFT.

\begin{table*}[h]
\caption{TTFT, throughput, and end-to-end latency at concurrency 1, comparing across serving systems under original settings.}
\label{tab:e2e_latency_bs1}
\centering
\begin{tabular}{ll|ccc}
\toprule
\textbf{Algorithm} & \textbf{Implementation}
& \textbf{Throughput/(token/s)} & \textbf{TTFT/s} & \textbf{Latency/s} \\
\midrule

\multirow{3}{*}{R2R}
& LLM-only
& 40.17
& 0.074
& 388.12 \\

& Official Code
& 54.77
& 0.12
& 330.69 \\

& TokenRouter
& 105.81
& 0.11
& \textbf{144.10} \\

\midrule

\multirow{3}{*}{CITER}
& LLM-only
& 34.40
& 0.055
& 2.42 \\

& Official Code
& 14.26
& 0.031
& 2.41 \\

& TokenRouter
& 50.63
& 0.031
& \textbf{0.36} \\

\midrule

\multirow{3}{*}{Co-LLM}
& LLM-only
& 36.06
& 0.049
& 12.66 \\

& Official Code
& 2.80
& 0.34
& 76.68 \\

& TokenRouter
& 28.57
& 0.068
& \textbf{7.60} \\

\midrule

\multirow{3}{*}{R-Stitch}
& LLM-only
& 40.82
& 0.072
& 341.84 \\

& Official Code
& N/A & N/A & N/A \\

& TokenRouter
& 55.78
& 0.102
& \textbf{119.29} \\

\bottomrule
\end{tabular}
\end{table*}

\subsection{Additional Ablation Study}
\label{sec:appendix/ablation}

\xhdr{Gain breakdown}
We break down the throughput gains of TokenRouter to quantify the contribution of each optimization.
As shown in Table~\ref{tab:gain_source}, enabling CUDA graphs improves the throughput by 1.88--2.12$\times$.
On top of these gains, asynchronous execution and delayed batching further improve throughput by 1.12$\times$ and 1.31$\times$, respectively, at concurrency 4.
These results demonstrate that TokenRouter's overall performance gain stems from the combined benefits of these optimizations.

\begin{table}[h]
    \centering
    \caption{Ablation study of different optimization components. Throughput is reported in token/s.}
    \label{tab:gain_source}
    \setlength{\tabcolsep}{3pt}
    \renewcommand{\arraystretch}{1.13}
    \begin{tabular}{lccc}
    \toprule
    \textbf{Method} & $N=8$ & $N=4$ & $N=1$ \\
    \midrule
    TokenRouter & \textbf{372.48} & \textbf{210.61} & \textbf{73.02} \\
    \midrule
    - Delayed batching & 296.86 & 161.26 & 73.02 \\
    - Async execution & 230.79 & 143.88 & 73.02 \\
    - Router CUDA graph & 228.10 & 142.33 & 72.62 \\
    - LLM CUDA graph & 164.08 & 95.55 & 51.22 \\
    - SLM CUDA graph ($\approx$ R2R) & 132.78 & 75.54 & 34.20 \\
    \midrule
    R2R official code & 134.89 & 73.27 & 36.17 \\
    \bottomrule
    \end{tabular}
\end{table}

\xhdr{Parallel strategy} 
Table~\ref{tab:parallelism} studies how tensor-parallelism (TP) configurations for the SLM and LLM affect throughput. 
Taking the two-GPU setting as an example, increasing the LLM TP size from 1 to 2 improves throughput by 40.6\% at $N=1$ and 22.8\% at $N=4$. In contrast, once the LLM uses TP size 2, further increasing the SLM TP size from 1 to 2 brings only 0.8\% and 0.4\% throughput improvements, respectively. This indicates that additional TP resources are substantially more beneficial to the LLM than to the SLM. Therefore, we prioritize tensor parallelism for the LLM and serve the SLM with TP size 1 by default throughout the paper.

\begin{table}[h]
    \centering
    \caption{Throughput across parallel strategies. Throughput is reported in token/s.}
    \label{tab:parallelism}
    \setlength{\tabcolsep}{3pt}
    \renewcommand{\arraystretch}{0.95}
    \begin{tabular}{ccccrr}
    \toprule
    \textbf{Total} & \textbf{\#SLM} & \textbf{\#LLM} & \textbf{Overlap} & \textbf{$N{=}1$} & \textbf{$N{=}4$} \\
    \midrule
    \multirow{3}{*}{2}
        & 1 & 1 & $\times$     & 50.37  & 160.74 \\
    \cmidrule(l){2-6}
        & 1 & 2 & $\checkmark$ & 70.83  & 197.40 \\
        & 2 & 2 & $\checkmark$ & 71.41  & 198.13 \\
    \midrule
    \multirow{5}{*}{8}
        & 4 & 4 & $\times$     & 93.23  & 284.62 \\
    \cmidrule(l){2-6}
        & 1 & 8 & $\checkmark$ & 109.67 & 296.34 \\
        & 2 & 8 & $\checkmark$ & 106.02 & 289.08 \\
        & 4 & 8 & $\checkmark$ & 110.02 & 297.63 \\
        & 8 & 8 & $\checkmark$ & 105.59 & 288.35 \\
    \bottomrule
    \end{tabular}
\end{table}

% \textbf{Effect of model pairs. } 
% Table~\ref{tab:r2r_tokenrouter_gain_app} evaluates \name against the most competitive R2R system baseline across diverse SLM-LLM pairs. For each pair, the SLM runs on one GPU, while the LLM runs on two GPUs using tensor parallelism. The SLM and LLM share the GPUs through CUDA MPS.
% The results show that \name consistently improves throughput across all model pairs. Compared with the official implementation of R2R, \name achieves 1.99--3.21$\times$ throughput improvement across different model pairs. These results suggest that \name's system-level optimizations are effective across different SLM sizes and LLM scales.

% \begin{table}[h]
%     \centering
%     \small
%     \caption{R2R vs.\ \name across SLM-LLM pairs. Throughput is reported in token/s.}
%     \label{tab:r2r_tokenrouter_gain_app}
%     \setlength{\tabcolsep}{4pt}
%     \renewcommand{\arraystretch}{1.05}
%     \begin{tabular}{llccc}
%     \toprule
%     \textbf{SLM} & \textbf{LLM} & \textbf{R2R} & \textbf{\name} & \textbf{Gain} \\
%     \midrule
%     0.6B & 8B  & 105.1  & 336.98 & 3.21$\times$ \\
%     0.6B & 32B & 77.08  & 197.40 & 2.56$\times$ \\
%     1.7B & 8B  & 103.37 & 285.41 & 2.76$\times$ \\
%     4B   & 8B  & 106.67 & 211.81 & 1.99$\times$ \\
%     \bottomrule
%     \end{tabular}
% \end{table}

\subsection{Additional Deployment Configurations}

\subsubsection{Non-overlapping GPU Deployment}

% In Section \ref{sec:experiment}, we deliberately align the total number of GPUs between the token-routing method and the LLM baseline. Such a setup explicitly provides an apples-to-apples comparison under the same computational resources.

We deploy Qwen3-0.6B (1 GPU) and Qwen3-32B (TP=2, 2 GPUs) on separate GPUs, so that the LLM in token-routing uses the same resources as the LLM-only baseline. This configuration improves the throughput of TokenRouter by 6.43\% on average compared with the original overlapping setting.
\begin{table}[h]
\centering
\caption{Throughput (token/s) comparison across routing strategies under different concurrency and overlap settings.}
\label{tab:overlap_comparison}

\setlength{\tabcolsep}{4pt}
\renewcommand{\arraystretch}{1.08}

\begin{tabular}{clcccc}
\toprule

\multicolumn{2}{c}{\textbf{Setup}} & 
\multicolumn{4}{c}{\textbf{Routing Algorithm}} \\

\cmidrule(lr){1-2}\cmidrule(lr){3-6}

\textbf{$N$} & \textbf{Deployment} &
\textbf{R2R} & \textbf{R-Stitch} & \textbf{Co-LLM} & \textbf{CITER} \\

\midrule

\multirow{2}{*}{1}
& Overlapping 
& 73.27 & 43.06 & 47.46 & 102.29 \\

& Non-overlapping
& 74.90 & 47.53 & 48.25 & 108.65 \\

\midrule

\multirow{2}{*}{4}
& Overlapping 
& 205.05 & 108.42 & 165.96 & 281.27 \\

& Non-overlapping
& 228.94 & 116.14 & 170.20 & 308.36 \\

\bottomrule

\end{tabular}
\end{table}

\subsubsection{Cross-node Deployment}

We evaluate TokenRouter under cross-node deployment setups. Specifically, we deploy Qwen3-0.6B (1 GPU) and Qwen3-32B (2 GPUs, TP=2) on one or two nodes and test the throughput under the low-effort reasoning setup. The two nodes are connected via RoCE at 3.7 GB/s. As shown in Table~\ref{tab:node_comparison}, cross-node throughput is comparable to single-node throughput with the same computational resources, translating to 1.93--32.81$\times$
 improvements over the respective official implementations.

\begin{table}[h]
\centering
\caption{Throughput comparison across single-node and multi-node settings.}
\label{tab:node_comparison}

\begin{tabular}{llccc}
\toprule

\multicolumn{2}{c}{\textbf{Setup}} & 
\multicolumn{3}{c}{\textbf{Routing Algorithm}} \\

\cmidrule(lr){1-2}\cmidrule(lr){3-5}

\textbf{$N$} & \textbf{Deployment} &
\textbf{R-Stitch} & \textbf{R2R} & \textbf{CITER} \\

\midrule

\multirow{2}{*}{1}
& Single-node
& 47.53 & 74.90 & \textbf{108.65} \\

& Multi-node
& 46.19 & 69.92 & \textbf{93.99} \\

\midrule

\multirow{2}{*}{4}
& Single-node
& 116.14 & 228.94 & \textbf{308.36} \\

& Multi-node
& 104.20 & 202.43 & \textbf{277.56} \\

\bottomrule

\end{tabular}

\end{table}

\subsection{Latency Breakdown of the Std. Serving Baseline}

To understand the gap over the CUDA-graph-enabled Std. Serving baseline, we profile the average SLM-side runtime per decoding step over the final 128 steps of a 2,048-token generation. As shown in Table~\ref{tab:slm_latency_breakdown}, the majority of latency is attributed to prefix matching and RadixCache maintenance. This overhead stems from the single-model assumption in Std. Serving, where each returning request is treated as a new extend operation from the user side, incurring repeated prefix matching and KV-cache bookkeeping. In contrast, TokenRouter preserves the request-level KV-cache state across model handoffs and directly resumes execution when peer results return, thereby substantially reducing such overhead.

\begin{table}[h]
\centering
\caption{Latency breakdown of the SLM-side execution per decoding step in Std. Serving.}
\label{tab:slm_latency_breakdown}

\setlength{\tabcolsep}{4pt}
\renewcommand{\arraystretch}{1.08}

\begin{tabular}{lcc}
\toprule

\textbf{Operation} & \textbf{Latency (ms)} & \textbf{Ratio} \\

\midrule

Prefix matching
& 7.78 & 20.94\% \\

Locking cache nodes
& 6.49 & 17.47\% \\

SLM inference
& 1.56 & 4.20\% \\

Releasing locks
& 5.13 & 13.81\% \\

Update radix cache
& 12.12 & 32.62\% \\

Others
& 4.08 & 10.98\% \\

\bottomrule

\end{tabular}

\end{table}

\section{Token-Routing-Specific Optimizations}
\label{appendix:opt}

Beyond the core mechanisms in Section~\ref{sec:architecture}, \name adapts several single-LLM serving assumptions that token-level routing breaks. Each subsection first identifies the broken assumption and then describes \name's fix.

\xhdr{Automatic KV-cache memory allocation}
Single-LLM servers~\citep{kwon2023vllm,zheng2024sglang} assume one model owns the entire KV-cache budget on the GPU. Token-level routing co-locates cooperating LLMs on the same GPU and forces them to contend for this budget. Over-reserving per model then causes out-of-memory at full concurrency, while under-reserving idles capacity.
\name resolves this by treating per-model KV reservations as a joint allocation. Given each model's weight footprint and per-token KV-cell size, the runtime computes the largest common KV-token capacity that fits within a configurable shared GPU budget and assigns each subserver its share before launch. This eliminates manual reservation tuning.

\xhdr{CUDA-graph capture for extend mode}
Single-LLM servers capture CUDA graphs only for decode batches, where the per-step shape is stable. Under token-level routing, however, every peer return arrives as an extend batch, so extend mode (rather than decode) dominates the routing-handoff path.
\name therefore captures CUDA graphs over a configurable range of extend batch sizes and token lengths and falls back to the standard non-graph extend path when a batch exceeds this range. Since the common case stays within the captured range, per-handoff kernel-launch overhead is eliminated on the dominant path while long extends remain correctly served.

\xhdr{In-process router}
In single-LLM and query-level serving, the router (when present) runs once per request, so a cross-process invocation is acceptable. Token-level routing inverts this scaling, since the \textbf{route} function runs after every forward pass on every active request, orders of magnitude more frequently. A cross-process router would add an IPC round-trip per token and serialize routing against the next decoding step.
\name therefore co-locates the router with the scheduler on the same GPU as the LLM runner, so each route decision becomes an in-process tensor op.

\xhdr{Global state tracking}
The delayed-batching threshold (Section~\ref{sec:architecture/delayed}) depends on the system-wide active request count, so all subservers must agree on this count. Otherwise $B^{*}$ would be selected against stale concurrency and drift from optimal.
Rather than adding a central coordinator, \name piggybacks status updates on the peer-message channel. Every newly admitted or finished request is announced as a control message alongside data tokens, keeping each subserver's view consistent without separate coordination infrastructure.

\section{Implementation Details}
\label{appendix:impl}

% \todo{not only for R2R; should be generalizable enough}

\name is implemented on top of SGLang~\citep{zheng2024sglang} 0.5.1. The Server Interface is adapted from SGLang's FastAPI front-end to support multi-model port arguments, and inter-subserver messages travel over local ZeroMQ IPC. Each subserver runs its three-stage iteration as a single event loop in a dedicated subprocess, and the only inter-subserver coordination is the peer-message channel of Section~\ref{sec:architecture/async}. All evaluated routing algorithms (Section~\ref{sec:experiment}) share this runtime, each implemented as a scheduler subclass over the three functions of Section~\ref{sec:programming}.

\section{Mathematical Modeling for Delayed-Batching Optimization}
\label{appendix:math_model}

\begin{table}[htbp]
    \centering
    \small
    \caption{Key notations used throughout the analysis.}
    \label{tab:general-routing-notation}
    \renewcommand{\arraystretch}{1.08}
    \begin{tabular}{p{0.10\textwidth}p{0.80\textwidth}}
        \toprule
        Notation & Definition \\
        \midrule
        $M$ & Number of cooperating models. \\
        $N$ & Number of concurrent requests in the system. \\
        $L_i$ & Latency of one decoding step on model $i$. \\
        $p_{ij}$ & Probability that a request is routed from model $i$ to model $j$ after a decoding step. \\
        $c_{ij}$ & Number of output tokens committed when a request is routed from model $i$ to model $j$. \\
        $B_i$ & Delayed-batching threshold of model $i$. \\
        $k_i$ & Number of requests waiting at model $i$. \\
        $b_i$ & Size of the currently executing batch on model $i$. \\
        $r_i$ & Residual execution time of the currently executing batch on model $i$. \\
        $\mathcal{R}_{\mathbf{B}}$ & Set of states reachable under delayed-batching threshold $\mathbf{B}$. \\
        $\tau(\mathbf{s})$ & Holding time of state $\mathbf{s}$. \\
        $\mathcal{A}(\mathbf{s})$ & whose current batches complete at the next state transition. \\
        $\mathbf{m_i}$ & Destination-count vector for the batch completed on model $i$. \\
        $\mathbf{T}_{\mathbf B},\boldsymbol\pi_{\mathbf B}$ & Transition matrix and stationary distribution of the DTMC model under $\mathbf B$. \\
        $g(\mathbf{s}, \mathbf{M})$ & Number of tokens committed during a state transition from $\mathbf{s}$ with routing outcome $\mathbf{M}$. \\
        $T(\mathbf{B})$ & Steady-state aggregate committed-token throughput under delayed-batching threshold $\mathbf{B}$. \\
        \bottomrule
    \end{tabular}
\end{table}

We develop a mathematical model to optimize the delayed-batching threshold introduced in Section~\ref{sec:architecture/delayed}.
We first formulate the problem in Section~\ref{appendix:formulation}, then derive the steady-state system throughput in Section~\ref{appendix:derivation} and optimize the batching threshold in Section~\ref{appendix:optimization}.
Finally, we use R2R~\citep{fu2025r2r} as a representative case study to validate the model in Section~\ref{appendix:example}.
The key notation used throughout the analysis is summarized in Table~\ref{tab:general-routing-notation}.

\subsection{Problem Formulation}
\label{appendix:formulation}
Consider a token-level routing system consisting of $M$ models, indexed by $i\in\{0,\ldots,M-1\}$, serving $N$ concurrent requests.
Let $L_i$ denote the latency of one decoding step on model $i$.

After completing a decoding step on model $i$, a request is routed to model $j$ with probability $p_{ij}$ and commits $c_{ij}$ output tokens.
We define the routing transition matrix and the token-commit matrix as
\begin{equation}
    \mathbf{P}=[p_{ij}]_{M\times M},\qquad \mathbf{C}=[c_{ij}]_{M\times M},
\end{equation}
where
\begin{equation}
    p_{ij}\ge 0,\qquad \sum_{j=0}^{M-1} p_{ij}=1.
\end{equation}
In particular, $p_{ii}$ denotes the probability that a request remains on model $i$ for the next decoding step.
We assume that the routing probabilities are stationary across decoding steps.
If the routing probabilities vary over time, the same analysis applies over a short time interval during which they can be approximated as constant.
Under this assumption, for $p_{ii}<1$, the number of consecutive decoding steps a request executes on model $i$ before transitioning to another model follows a geometric distribution with parameter $1-p_{ii}$.
We further assume that routing decisions are independent across requests.

To improve efficiency, each model $i$ employs delayed batching with threshold $B_i$: a new batch is launched only when at least $B_i$ requests are available in its waiting queue.
We denote the threshold vector by
\begin{equation}
    \mathbf{B}=(B_0,\ldots,B_{M-1}).
\end{equation}

Our objective is to characterize the steady-state throughput of \name as a function of $\mathbf{B}$ and determine the throughput-optimal threshold vector.

\subsection{Derivation of System Throughput}
\label{appendix:derivation}

\subsubsection{Problem Modeling: Discrete-Time Markov Chain}
We represent the system state as
\begin{equation}
    \mathbf{s}:=(k_0,b_0,r_0,\dots,k_{M-1},b_{M-1},r_{M-1}),
\end{equation}
where $k_i$ is the number of requests waiting at model $i$, $b_i$ is the size of model $i$'s currently running batch, and $r_i$ the residual execution time of that batch, measured immediately after the most recent batch-completion event in the system.
If model $i$ has no running batch, we set $b_i=r_i=0$.
Under a delayed-batching threshold vector $\mathbf B$, the state space is
\begin{equation}
\begin{aligned}
    \mathcal{S}_{\mathbf B}
    :=\bigg\{\mathbf s\in\mathbb R^{3M}:&
    \sum_{i=0}^{M-1}(k_i+b_i)=N,\ k_i\in\mathbb{N},\ b_i\in\mathbb{N},\\
    &b_i=0\Rightarrow k_i<B_i,\ r_i=0,\\
    &b_i>0\Rightarrow b_i\ge B_i,\ 0<r_i\le L_i
    \bigg\}.
\end{aligned}
\end{equation}

Since the residual execution times are continuous, $\mathcal{S}_{\mathbf{B}}$ is generally infinite.
To obtain a finite-state model, we introduce a common time quantum $\delta$ and assume that the decoding-step latency of each model is an integer multiple of $\delta$:
\begin{equation}
    L_i=d_i\delta,\qquad d_i\in\mathbb{N},~i=0,1,\dots,M-1.
\end{equation}
We further assume that all initial residual times are integer multiples of $\delta$.
Then all subsequent residual times remain on the same discrete grid, yielding the finite candidate state space
\begin{equation}
\begin{aligned}
    \tilde{\mathcal{S}}_{\mathbf B}
    :=\bigg\{\mathbf s\in\mathbb R^{3M}:&
    \sum_{i=0}^{M-1}(k_i+b_i)=N,\ k_i\in\mathbb{N},\ b_i\in\mathbb{N},\\
    &b_i=0\Rightarrow k_i<B_i,\ r_i=0,\\
    &b_i>0\Rightarrow b_i\ge B_i,\ r_i\in\{\delta,2\delta,\dots,d_i\delta\}
    \bigg\}.
\end{aligned}
\end{equation}

We define a state transition at each batch-completion event, i.e., whenever one or more models complete their current batches.
Because the state records the queue length, running batch size, and residual execution time of every model, the distribution of the next state depends only on the current state.
Therefore, the system observed at batch-completion epochs forms a finite-state discrete-time Markov chain (DTMC).

For a fixed $\mathbf{B}$ and an initial state $\mathbf{s}_0\in\tilde{\mathcal{S}}_{\mathbf{B}}$, let $\mathcal{R}_{\mathbf B}$ denote the set of states reachable from $\mathbf{s}0$ with positive probability under the routing and scheduling dynamics.
In practice, $\mathcal{R}_{\mathbf B}$ can be enumerated using breadth-first search (BFS).
By construction, $\mathcal{R}_{\mathbf B}\subseteq\tilde{\mathcal{S}}_{\mathbf B}$ and is therefore finite.

\subsubsection{Transition Matrix and Stationary Distribution}
We next characterize the state-transition dynamics of the DTMC and derive its stationary distribution, which will be used to compute the steady-state system throughput.

For a state $\mathbf{s}$, define its holding time as
\begin{equation}
    \tau(\mathbf{s}):=\min_{i:b_i>0}r_i.
\end{equation}
The set of models whose current batches complete at the next transition is therefore
\begin{equation}
    \mathcal{A}(\mathbf{s}):=\{i:b_i>0,~r_i=\tau(\mathbf{s})\}.
\end{equation}
For each model $i\in\mathcal{A}(\mathbf{s})$, its batch of $b_i$ requests completes one decoding step and is subsequently routed to the next models.
Let $\mathbf{m}_i=(m_{i0},\dots,m_{i,M-1})$ be the corresponding destination-count vector, where $m_{ij}$ is the number of requests routed from model $i$ to model $j$.
By definition,
\begin{equation}
    \sum_{j=0}^{M-1}m_{ij}=b_i.
\end{equation}
Conditioned on $\mathbf{s}$, $\mathbf{m}_i$ follows a multinomial distribution:
\begin{equation}
    P_i(\mathbf m_i\mid \mathbf{s})
    =
    \frac{b_i!}{\prod_{j=0}^{M-1}m_{ij}!}
    \prod_{j=0}^{M-1}p_{ij}^{m_{ij}},
\end{equation}
with
\begin{equation}
    \mathbb{E}[m_{ij}\mid \mathbf{s}]=p_{ij}b_i.
\label{eq:multinomial_expectation}
\end{equation}

We collect the routing outcomes of all completing models sas $\mathbf{M}(\mathbf{s})=(\mathbf{m}_i)_{i\in\mathcal{A}(\mathbf{s})}$.
Under the independence assumption across requests, their joint probability is
\begin{equation}
    P(\mathbf{M}\mid\mathbf{s})=\prod_{i\in\mathcal A(\mathbf s)}P_i(\mathbf m_i\mid \mathbf{s}).
\end{equation}

After the completed requests are routed and all batches that satisfy their delayed-batching thresholds are launched, the system transitions to
\begin{equation}
    \mathbf{s}'=\Phi(\mathbf{s},\mathbf{M},\mathbf{B}),
\end{equation}
where $\Phi$ denotes the deterministic state-update function induced by the routing outcome and the delayed-batching policy.

We define the transition matrix $\mathbf{T}_{\mathbf{B}}$ over the reachable state space $\mathcal{R}_{\mathbf B}$. 
For any $\mathbf{s},\mathbf{s}'\in\mathcal{R}_{\mathbf B}$,
\begin{equation}
    \mathbf{T}_{\mathbf{B}}(\mathbf{s},\mathbf{s}')
    =P(\mathbf{s}_{t+1}=\mathbf{s}'\mid \mathbf{s}_t=\mathbf{s})
    =\sum_{\mathbf{M}:\Phi(\mathbf{s},\mathbf{M},\mathbf{B})=\mathbf{s}'}P(\mathbf{M}\mid\mathbf{s}).
\end{equation}
If the finite-state chain on $\mathcal{R}_{\mathbf B}$ is irreducible, it has a unique stationary distribution $\boldsymbol{\pi}_{\mathbf{B}}$ satisfying
\begin{equation}
    (\mathbf{T}_{\mathbf{B}}^{\top}-\mathbf{I})\boldsymbol{\pi}_{\mathbf{B}}=\mathbf{0},
    \qquad
    \sum_{\mathbf{s}\in\mathcal{R}_{\mathbf{B}}}\pi_{\mathbf{B}}(\mathbf{s})=1.
\label{eq:stationary}
\end{equation}
Empirically, we observe that the reachable-state chain is irreducible for all configurations encountered during optimization.

\subsubsection{Throughput Derivation}
We now derive the steady-state aggregate committed-token throughput from the stationary distribution of the DTMC.

For a transition from state $\mathbf{s}$ with routing outcome $\mathbf{M}$, the number of committed tokens is
\begin{equation}
    g(\mathbf{s},\mathbf{M})
    :=\sum_{i\in\mathcal{A}(\mathbf{s})}\sum_{j=0}^{M-1}c_{ij}m_{ij}.
\label{eq:committed_tokens}
\end{equation}
The long-run aggregate committed-token throughput is therefore
\begin{equation}
\begin{aligned}
    T(\mathbf B)
    &=\lim_{H\to\infty}
    \frac{\displaystyle\sum_{n=0}^{H-1}g(\mathbf{s}_n,\mathbf{M}_n)}
    {\displaystyle\sum_{n=0}^{H-1}\tau(\mathbf{s}_n)}\\
    &=\frac{\displaystyle\lim_{H\to\infty}\frac1H\sum_{n=0}^{H-1}g(\mathbf{s}_n,\mathbf{M}_n)}
    {\displaystyle\lim_{H\to\infty}\frac1H\sum_{n=0}^{H-1}\tau(\mathbf{s}_n)}\\
    &=\frac{\mathbb{E}[g(\mathbf{s},\mathbf{M})]}
    {\mathbb{E}[\tau(\mathbf{s})]}.
\label{eq:throughput_expectation}
\end{aligned}
\end{equation}

We first derive the expected number of committed tokens per state transition.
Taking the expectation of Equation~\ref{eq:committed_tokens} and applying Equation~\ref{eq:multinomial_expectation}, we obtain
\begin{equation}
\begin{aligned}
    \mathbb{E}[g(\mathbf{s},\mathbf{M})]
    &=\mathbb{E}_{\mathbf{s}}\!\left[\mathbb{E}_{\mathbf{M}}[g(\mathbf{s},\mathbf{M})\mid \mathbf{s}]\right]\\
    &=\mathbb{E}_{\mathbf{s}}\!\left[
        \sum_{i\in\mathcal{A}(\mathbf{s})}\sum_{j=0}^{M-1}
        c_{ij}\mathbb{E}_{\mathbf{M}}[m_{ij}\mid \mathbf{s}]
      \right]\\
    &=\mathbb{E}_{\mathbf{s}}\!\left[
        \sum_{i\in\mathcal{A}(\mathbf{s})}\sum_{j=0}^{M-1}
        b_ic_{ij}p_{ij}
      \right]\\
    &=\sum_{\mathbf{s}\in\mathcal{R}_\mathbf{B}}
      \sum_{i\in\mathcal{A}(\mathbf{s})}\sum_{j=0}^{M-1}
      \pi_{\mathbf{B}}(\mathbf{s})b_ic_{ij}p_{ij}.
\label{eq:token_expectation}
\end{aligned}
\end{equation}
Similarly, the expected time between consecutive state transitions is
\begin{equation}
    \mathbb{E}[\tau(\mathbf{s})]
    =\sum_{\mathbf{s}\in\mathcal{R}_\mathbf{B}}
    \pi_{\mathbf{B}}(\mathbf{s})\tau(\mathbf{s}).
\label{eq:time_expectation}
\end{equation}
Substituting Equation~\ref{eq:token_expectation} and~\ref{eq:time_expectation} into Equation~\ref{eq:throughput_expectation} gives
\begin{equation}
    T(\mathbf B)
    =\frac{
        \displaystyle\sum_{\mathbf{s}\in\mathcal{R}_\mathbf{B}}
        \sum_{i\in\mathcal{A}(\mathbf{s})}\sum_{j=0}^{M-1}
        \pi_{\mathbf{B}}(\mathbf{s})b_ic_{ij}p_{ij}}
      {\displaystyle\sum_{\mathbf{s}\in\mathcal{R}_\mathbf{B}}
        \pi_{\mathbf{B}}(\mathbf{s})\tau(\mathbf{s})}.
\label{eq:throughput}
\end{equation}

\subsection{Delayed-Batching Threshold Optimization}
\label{appendix:optimization}
We now optimize the delayed-batching threshold vector $\mathbf{B}$ using the throughput model derived above.

To avoid deadlock states in which no model has enough queued requests to launch a batch, we impose the feasibility condition
\begin{equation}
    \sum_{i=0}^{M-1}(B_i-1)<N.
    \label{eq:general-safe-condition}
\end{equation}
This condition defines the finite feasible set
\begin{equation}
    \mathcal B_M
    :=
    \left\{
        \mathbf B\in\{1,\dots,N\}^M:
        \ \sum_{i=0}^{M-1}(B_i-1)<N
    \right\}.
    \label{eq:general-safe-domain}
\end{equation}
The throughput-optimal delayed-batching threshold vector is then given by
\begin{equation}
    \mathbf B^\star
    =
    \operatorname*{arg\,max}_{\mathbf B\in\mathcal B_M}
    \frac{
        \displaystyle\sum_{\mathbf{s}\in\mathcal{R}_\mathbf{B}}
        \sum_{i\in\mathcal{A}(\mathbf{s})}\sum_{j=0}^{M-1}
        \pi_{\mathbf{B}}(\mathbf{s})b_ic_{ij}p_{ij}}
      {\displaystyle\sum_{\mathbf{s}\in\mathcal{R}_\mathbf{B}}
        \pi_{\mathbf{B}}(\mathbf{s})\min_{i:b_i>0}r_i},
\end{equation}
where $\boldsymbol{\pi}_{\mathbf{B}}$is the stationary distribution obtained by solving Equation~\ref{eq:stationary}.

\subsection{Empirical Validation of the Throughput Model}
\label{appendix:example}

\begin{figure}[t]
    \centering
    \includegraphics[width=1\linewidth]{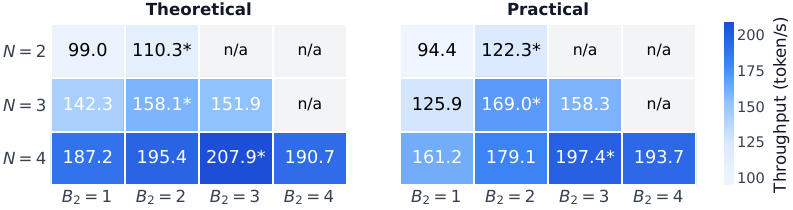}
    \caption{Predicted and measured R2R throughput under different concurrency levels and delayed-batching thresholds. We fix the SLM threshold at $B_0=1$ and vary the LLM threshold $B_1$. An asterisk (*) marks the highest throughput for each concurrency level in each panel.}
    \label{fig:delayed_batching}
\end{figure}

We empirically validate the throughput model using R2R~\cite{fu2025r2r}, a representative token-level routing algorithm with $M=2$ models.
Following the indexing convention used throughout our formulation, model 0 denotes the SLM and model 1 the LLM.
For R2R, the routing-probability matrix $\mathbf P$ and token-commit matrix $\mathbf C$ are given by
\begin{equation}
    \mathbf P
    =
    \begin{bmatrix}
        1-p & p\\
        1 & 0
    \end{bmatrix},
    \qquad
    \mathbf C
    =
    \begin{bmatrix}
        1 & 0\\
        1 & 0
    \end{bmatrix}.
\end{equation}
Under the experimental setup described in Section~\ref{sec:exp/setup}, $p=0.35$.
The measured decoding-step latencies of two models are $L_0=6.0\,\mathrm{ms}$ for the SLM and $L_1=27.9\,\mathrm{ms}$ for the LLM. 
We set $\delta=0.1\,\mathrm{ms}$ and use Equation~\ref{eq:throughput} to predict the throughput under different concurrency levels $N$ and delayed-batching threshold vectors $\mathbf{B}$.

Figure~\ref{fig:delayed_batching} fixes $B_0=1$ and compares the predicted throughput against measurements obtained on hardware while varying $B_1$.
The model closely captures the throughput trends across both concurrency levels and batching thresholds.
More importantly, at every evaluated concurrency level, the threshold predicted to maximize throughput coincides with the empirically optimal threshold.
These results show that the proposed model accurately captures the effect of delayed batching on system throughput and can guide the selection of throughput-optimal batching thresholds.

\newpage

\end{document}